\documentclass[conference]{IEEEtran}\IEEEoverridecommandlockouts
\IEEEoverridecommandlockouts

\ifCLASSINFOpdf
  \usepackage[pdftex]{graphicx}
  \graphicspath{{fig/}}
\else
  \usepackage[dvips]{graphicx}
  \graphicspath{{fig/}}
  \DeclareGraphicsExtensions{.eps}
\fi
\usepackage[cmex10]{amsmath}
\usepackage{amssymb}
\newcommand{\argmax}{\operatornamewithlimits{argmax}}

\newcommand{\bvec}[1]{\mbox{\boldmath $#1$}}

\newcommand{\defeq}{\overset{\mathrm{def}}{=}}

\ifCLASSOPTIONcompsoc
 \usepackage[caption=false,font=normalsize,labelfont=sf,textfont=sf]{subfig}
\else
 \usepackage[caption=false,font=footnotesize]{subfig}
\fi
\usepackage{multirow}
\usepackage{rotating}

\begin{document}
%\title{End-to-End Line-level Script Identification \\ for Multilingual OCR}
\title{Sequence-to-Label Script Identification \\ for Multilingual OCR}


\author{\IEEEauthorblockN{Yasuhisa Fujii\IEEEauthorrefmark{1}\thanks{\IEEEauthorrefmark{1}: These authors contributed equally.}, Karel Driesen\IEEEauthorrefmark{1}, Jonathan Baccash,
  Ash Hurst, Ashok C. Popat}
\IEEEauthorblockA{Google Research, Mountain View, CA 94043, USA\\
  \{yasuhisaf,karel,jbaccash,phurst,popat\}@google.com} }


%Telephone: (800) 555--1212, Fax: (888) 555--1212}


\maketitle

\begin{abstract}
  We describe a novel line-level script identification method.
  Previous work
  repurposed an OCR model generating per-character script codes,
  counted to obtain line-level script identification.
  This has two shortcomings. First, as a
  \emph{sequence-to-sequence} model it is more complex than necessary for the
  \emph{sequence-to-label} problem of line script identification. This
  makes it harder
  to train and inefficient to run. Second, the counting heuristic may
  be suboptimal compared to a learned model. Therefore we reframe line
  script identification as a sequence-to-label problem and
  solve it using two components, trained end-to-end: \emph{Encoder}
  and \emph{Summarizer}. The encoder converts a line image into a
  feature sequence. The summarizer aggregates the sequence to
  classify the line. We test various summarizers with
  identical inception-style convolutional networks as
  encoders. Experiments on scanned books and photos containing 232
  languages in 30 scripts show 16\% reduction of script identification
  error rate compared to the baseline. This improved script
  identification reduces the character error rate attributable
  to script misidentification by 33\%.
\end{abstract}

% We achieved X\% and Y\% improvement over the baseline in terms of script
% identification and end-to-end character error rates, respectvely, on scanned
% books dataset.  We achieved X\% and Y\% improvement over the baseline in terms
% of script identification and end-to-end character error rates, respectively, on
% more challenging web and photo dataset.


% \fi
\IEEEpeerreviewmaketitle

\section{Introduction}

% Optical character recognition (OCR) is a classical, fundamental, and unsolved
% research problem. In the past decades, great progress has been made in this
% field~\cite{H-Wada:1959:UNESCO,Henry-S-Baird:1992:IEEE,GeorgeNagy:2000:TPAMI}.
% Contemporary studies on character recognition deal with challenging historical
% documents~\cite{} and images in unconstrained conditions such as photos and
% incidentals~\cite{AlessandroBissaco:2013:ICCV,ICDAR2015-RobustReading}, as well
% as modern scanned document and book images.

% Yet high quality OCR is still not readily available for many languages.  Even
% when a system is available, it is often specialized to that language or language
% group, employing interfaces and conventions specific to the vendor.  This poses
% a challenge in situations where it is desired from the outset to handle material
% in a large number of languages, as may be the case in a large-scale book
% scanning effort.

Multilingual optical character recognition (OCR), where the scripts and
languages of input images are unknown, is a challenging
problem in OCR. This holds in particular for multi-script images.
Most existing systems assume scripts to be known in advance and
do not handle mixed or misspecified scripts gracefully.
Ideally, OCR should work fast and accurately with or without script
specification.

A straightforward approach to handle arbitrary scripts and languages is to cover
all of them by one model.  The resulting mammoth system recognizes text in a
broad range of scripts, but this universality can cost accuracy and speed.
Alternatively, we can model each individual script or language and use model
selection as a first stage. Genzel et al. presented a multilingual OCR system
based on line-level script identification where each script and language has its
own OCR model~\cite{DmitriyGenzel:2013:HSI:2505377.2505382}. We adopt a similar
approach. However, we use a coarser, script-based model taxonomy where a model
recognizes any language in a given script. Section~\ref{sec:multilingual-ocr}
describes our multilingual OCR system. Note that selection is done at the
text-line level, i.e. each line is assumed to be in a single script.
Section~\ref{sec:discussion} discusses this choice. We use script identification
for OCR model selection. Therefore, it is a crucial component of the system.

% For example, in this paper, we deal with 30 scripts and 232
% languages, which contain $\sim$40000 different characters (more precisely atoms
% as explained in Section~\ref{sec:multilingual-character-processing}). Such a
% large fanout leads to many opportunities for symbol confusion and the system
% which we in fact trained had inferior accuracy.

% Script identification is a well-studied topic. Earlier efforts focussed
% mainly on scanned documents with white background and excellent resolution
% ~\cite{Ghosh:2010:SRR:1907655.1908052}
% ~\cite{Pati:2008:WLM:1371261.1371355}
% ~\cite{Chanda:2009:TAW:1634930.1635516}.
% More recent efforts deal with the messier environment of video and still
% camera images
% ~\cite{DBLP:conf/icdar/SharmaMSPB15}
% ~\cite{conf/das/SharmaPB12}
% ~\cite{conf/das/ZhaoSLT12}
% ~\cite{DBLP:conf/icdar/PhanSDLT11}.

Script identification as a \emph{pixels-to-label} problem is a well-studied
topic. Early efforts focused on scanned documents with white background and
excellent resolution~\cite{Ghosh:2010:SRR:1907655.1908052}. Recent efforts deal
with messier video and still camera
images~\cite{conf/das/SharmaPB12,BaoguangShi:2015:ICDAR}.

We perform script identification on an entire text line.
At the line level we can exploit the sequential nature of written text, similar
to speech recognition.
\cite{DmitriyGenzel:2013:HSI:2505377.2505382,AjeetKumarSingh:2015:ICDAR} use
sequence models for line-level script identification by framing the problem as a
sequence-to-sequence problem. In contrast, we frame the problem as a
conceptually simpler \emph{sequence-to-label} problem. We propose to model it
with two components trained end-to-end.  The first component is an
\emph{Encoder} to convert a line image to a sequence of features.  The second
component is a \emph{Summarizer} to take the feature sequence and produce a
single classification.  Similar formulations appear
in~\cite{BaoguangShi:2015:ICDAR} for script identification of scene texts and
in~\cite{KarelVesely:2016:ICASSP,GeorgHeigold:2016:ICASSP} for speaker
identification and verification in speech processing.  We design and evaluate
various summarizers with an inception-style~\cite{ChristianSzegedy:2015:CVPR}
convolutional network-based encoder in Section~\ref{sec:scriptid}.

% The contributions of the paper are two fold:
% \begin{itemize}
% \item We describe a multilingual OCR system supporting 232 languages in 30
%   scripts. To our knowledge, this is the first system with the coverage.
% \item We propose a novel line-level script identification method based on
%   encoder and summarizer. We investigate various summarizers and show that the
%   summarizer with a gating mechanism is superior to other approaches and
%   outperforms the strong baseline
%   from~\cite{DmitriyGenzel:2013:HSI:2505377.2505382} in terms of both script
%   identification and end-to-end character recognition with the multilingual OCR
%   system.
% \end{itemize}

\section{Multilingual OCR}
\label{sec:multilingual-ocr}

\subsection{Overview}
\label{sec:overview}

\begin{figure*}[!t]
\centering
\includegraphics[width=6.5in]{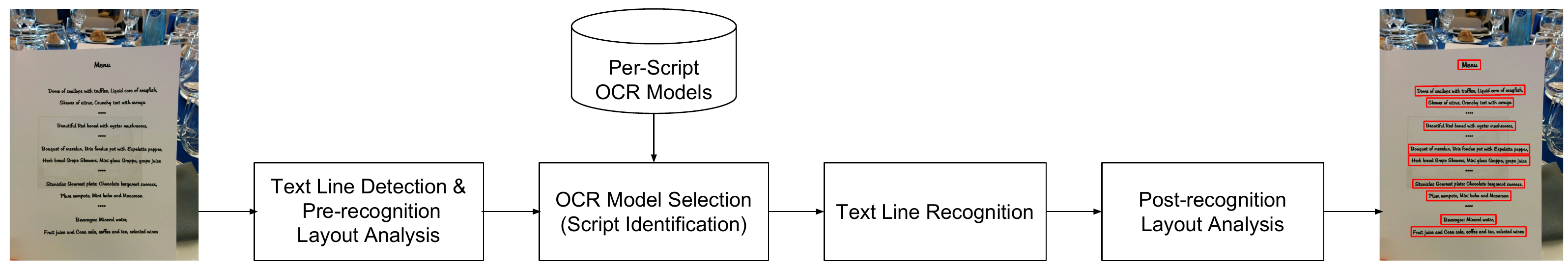}
\caption{Multilingual OCR system with line-level script identification.}
\label{fig:multilingual-ocr}
\end{figure*}

Fig.~\ref{fig:multilingual-ocr} shows our multilingual
OCR system.  The first step extracts text lines from the image.
This step
may provide structural layout information on paragraphs or sections.
It can also provide stylistic information such as bold or italic style,
as described
in~\cite{RaySmith:2009:ICDAR}.  Next, script identification selects a
script-specific OCR model among 30 scripts.
See Table~\ref{tab:data-description}. The script-specific OCR model then
produces a unicode transcription of the
line image. Finally, the recognition results further enhance
layout analysis (Post-recognition Layout Analysis). For example,
mistakes in the initial layout are
corrected post OCR by using a low recognition confidence
as a signal that
an initial text box was not actually text.

We focus on script identification and text line recognition, treating
text line detection as given.
Section~\ref{sec:scriptid} describes the details of script
identification methods.

\subsection{Text Line Recognition with Script Identification}
\label{sec:text-line-recognition-with-scriptid}

% A very straightforward way to build a multilingual OCR system is to train a
% single multi-script OCR model on all the script and language data. However, we
% do not take that approach in this paper since preliminary experiments showed
% that dealing with 30 scripts and 232 languages, which contain different
% $\sim$40000 atoms in total, by one model posed a different challenge. Instead,
% w We model the conditional probability for each script.

Text line recognition takes a text line image as input and produces a sequence
of Unicode code points, for example represented as UTF-8 strings. Let $\bvec{x}$
be an input line image and $\bvec{y}$ be a sequence of code points. We employ a
probabilistic approach to model the relationship and consider the conditional
probability $P(\bvec{y}|\bvec{x})$. Let $s \in S$ be a script. The script
information is incorporated into $P(\bvec{y}|\bvec{\bvec{x}})$ by considering
$s$ as a hidden variable. We compute $P(\bvec{y}|\bvec{x})$ as follows:
\begin{align}
  \label{eq:ocr-with-scriptid}
  P(\bvec{y}|\bvec{x}) &= \sum_{s}{P(\bvec{y},s|\bvec{x})} \notag \\
  &= \sum_s{P(\bvec{y}|s, \bvec{x})P(s|\bvec{x})} \notag \\
  &\approx P(\bvec{y}|\argmax_{s}{P(s|\bvec{x})}, \bvec{x}).
\end{align}
$P(\bvec{y}|s, \bvec{x})$ represents an OCR model of a script
$s$. $P(s|\bvec{x})$ represents a script identifier, which is described in
Section~\ref{sec:scriptid}. Note that the formulation employs hard
classification and only considers the script which maximizes
$P(s|\bvec{x})$ in text recognition. The
limitation of this approach is discussed in Section~\ref{sec:discussion}.

We use the following approximation to compute $P(\bvec{y}|s, \bvec{x})$:
\begin{align}
  \label{eq:cond-prob}
  P(\bvec{y}|s, \bvec{x}) &= \frac{P(\bvec{y}|s, \bvec{x})}{P(\bvec{y}|s)}P(\bvec{y}|s) \notag \\
  &\approx \prod_{i=1}^{|C(\bvec{y})|}\frac{P(c_i|s, \bvec{x})}{P(c_i|s)}P(\bvec{y}|s)
\end{align}
where $C(\bvec{y})$ is a function to convert $\bvec{y}$ to a corresponding
sequence of grapheme clusters $(c_1, c_2, \dots, c_{|C(\bvec{y})|})$, and where $c_i$
is the $i$-th grapheme cluster of $\bvec{y}$. $P(c|s, \bvec{x})$, $P(c|s)$ and
$P(\bvec{y}|s)$ are an optical model, a prior probability of a grapheme cluster
$c$, and a language model, respectively, for a script $s$. In practice, we
combine these probabilities by taking the logarithms and using the values as the
scores in a log-linear model. We use a log-linear model framework as
in~\cite{YasuhisaFujii:2015:ICDAR}.

\begin{figure}[!t]
\centering
\includegraphics[width=7cm]{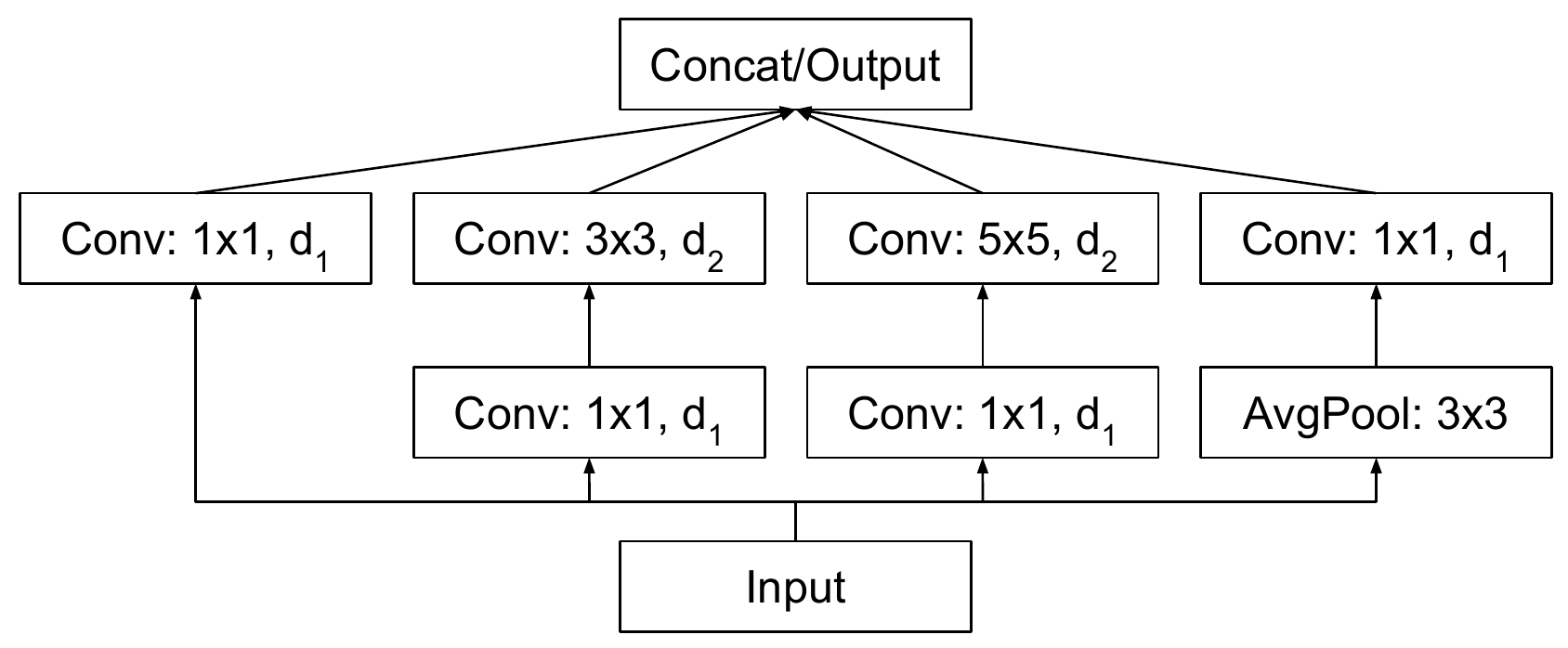}
\caption{Inception module used in the paper. ``Conv: $h_f{\times}w_f, d$''
  represents a convolutional neural network layer where $h_f$, $w_f$, and $d$
  mean the height of filter, the width of filter and the number of filters,
  respectively. The activation function is $\text{relu}6(x) = \min(\max(0, x),
  6)$. ``AvgPool: $h_f{\times}w_f$'' represents an average pooling layer. The
  stride of filters is always $1\times1$ for both layers.
  ``Concat'' concatenates
  all the inputs along the last dimension. The SAME padding is used for all
  operations.}
\label{fig:inception-module}
\end{figure}

\begin{figure}[!t]
\centering
\includegraphics[width=5cm]{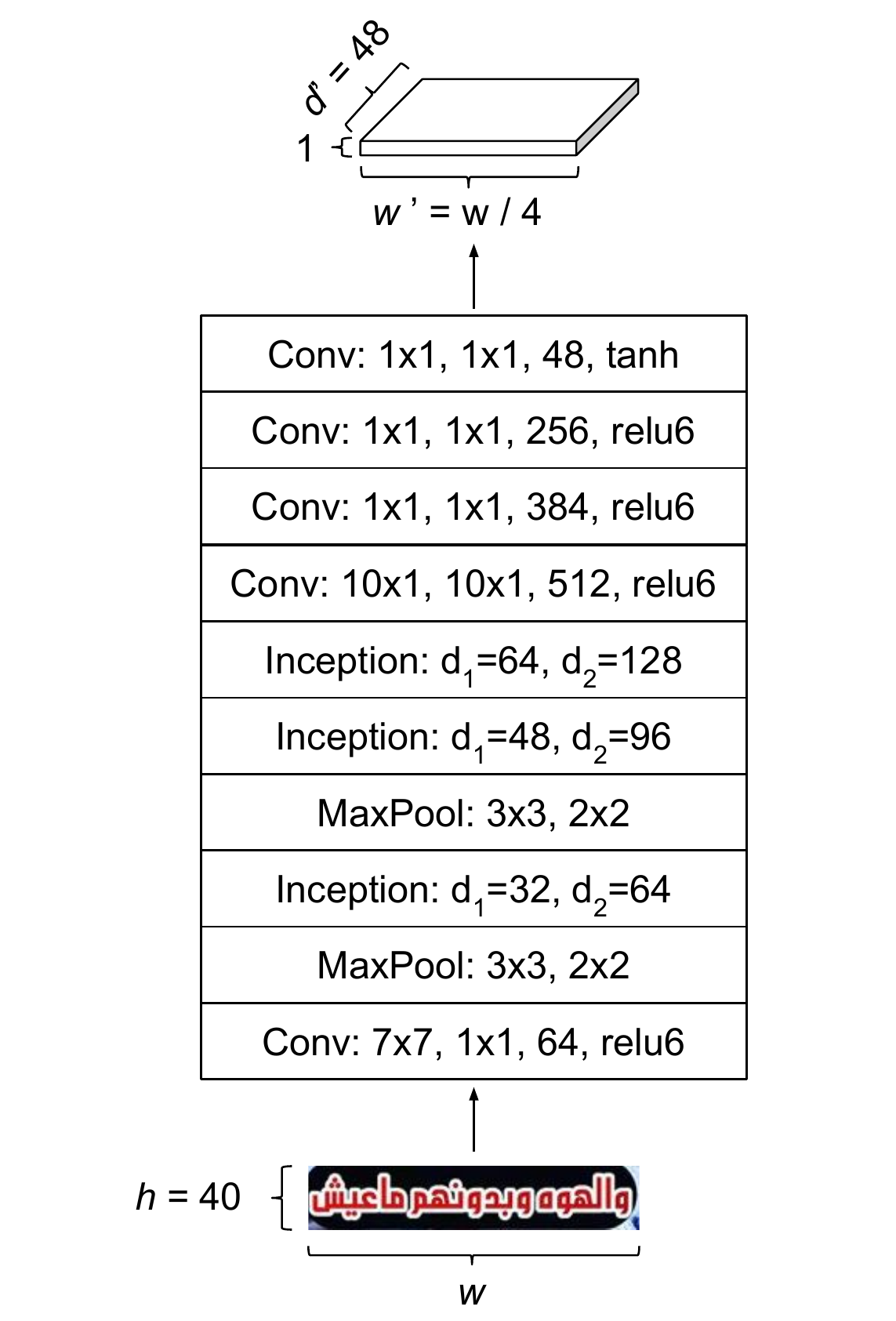}
\caption{Topology of the optical model. ``Conv: $h_f{\times}w_f, h_s{\times}w_s, c,
  f$'' represents a convolutional neural network layer where $h_f$, $w_f$,
  $h_s$, $w_s$, $c$ and $f$ mean the height of filter, the width of filter, the
  shift in the hight dimension, the shift in the width dimension and the number
  of filters and the activation function, respectively. $\text{tanh}(\bvec{x}) =
  (1 - \exp(-2x)) / (1 + \exp(-2x))$. ``MaxPool: $h_f{\times}w_f,
  h_s{\times}w_s$'' represents a max pooling layer. ``Inception: $d_1, d_2$''
  represents an inception module that is defined in
  Fig.~\ref{fig:inception-module}. The SAME padding is used for all operations.}
\label{fig:convolutional-layer}
\end{figure}

We model $P(c|s, \bvec{x})$ with a neural network. We assume that $\bvec{x}$ is
a tensor representing an image of fixed-height and has shape $(h, w, d)$ where
$h$ represents the fixed-height, $w$ is the width and $d$ is the number of
channels of the image. We fix the height to 40 and collapse the color space down to luminance (grayscale).
We use an inception-style convolutional neural
network as in ~\cite{ChristianSzegedy:2015:CVPR} to
convert $\bvec{x}$ into a tensor
representing a sequence of frames of shape $(w', d')$ where $w'$ is the length
of the sequence and $d'$ is the dimensionality of
features, see Fig.~\ref{fig:inception-module}.
Fig.~\ref{fig:convolutional-layer} shows the complete topology of our
model. Connectionist temporal classification (CTC)~\cite{AlexGraves2006ICML}
computes $\prod_{i}^{|C(\bvec{y})|}{P(c_i|s, \bvec{x})}$ from the output.

We use different models for horizontal and vertical texts of the same
script.  To deal with both directions uniformly we rotate vertical
texts 90 degrees counterclockwise and make them visually
horizontal. See Fig.~\ref{fig:text-line-images} for an example.
Right-to-left languages (e.g. Hebrew, Arabic) do not need special
handling since text line recognition deals with a line image as a
whole. Frames are offered to the network in display order.

$P(c|s)$ is the unigram language model of grapheme clusters and mathematically
needed to use the optical model $P(c|s, \bvec{x})$ with the language model
$P(\bvec{y}|s)$~\cite{Bourlard:1993:CSR:562393}. We estimate the probability by
computing the moving average of $P(c|s, \bvec{x})$ during training.

$P(\bvec{y}|s)$ is modeled as an $N$-gram language model defined over Unicode
points. We use (with pruning) $N=9$ for all scripts except Chinese, Japanese and
Korean for which we use $N=4$.  Note that the script $s$ may be used to
represent multiple languages. Therefore ``language model'' in the
foregoing should be understood in the functional sense of capturing
the known prior statistical regularity of the strings to be recognized,
rather than referring to a single actual language.

\section{Line-level Script Identification}
\label{sec:scriptid}

\begin{figure}[!t]
  \centering
  \includegraphics[width=4cm]{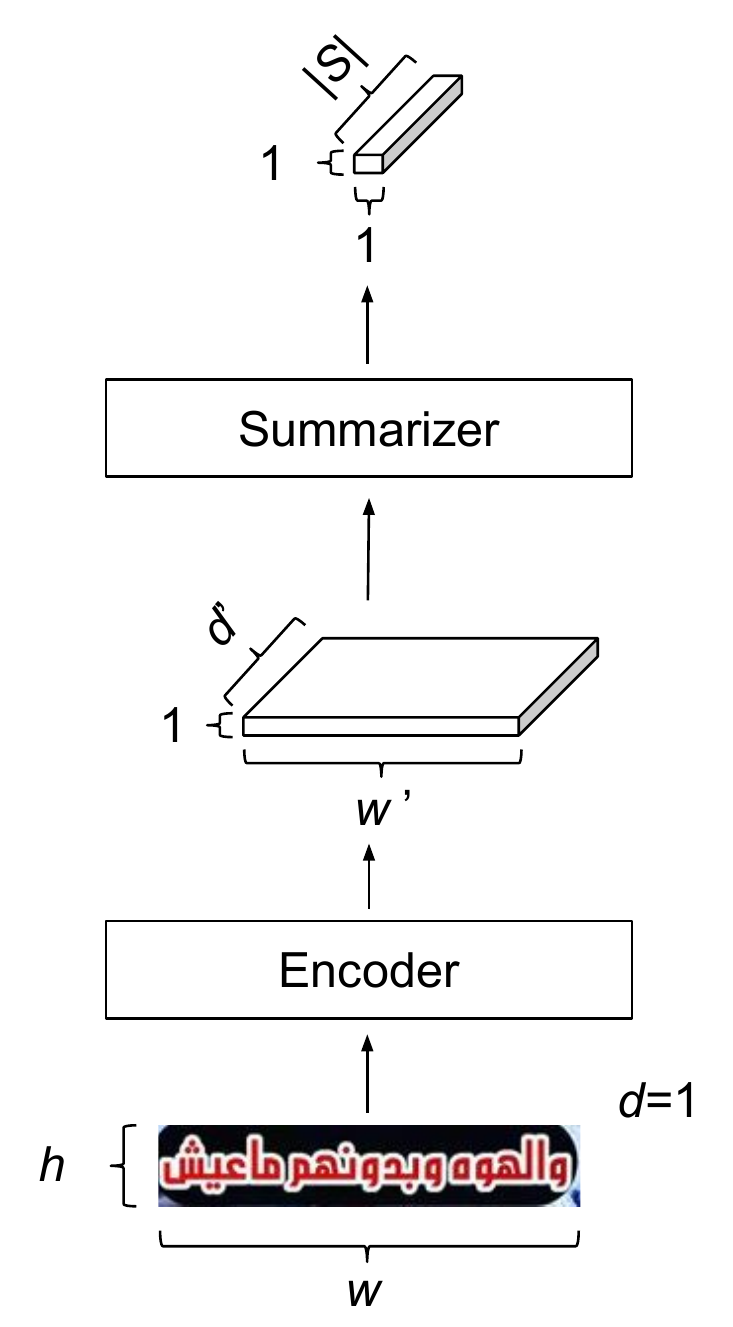}
  \caption{The encoder-summarizer framework to solve script identification as a
    sequence-to-label problem. An encoder processes an input of shape $(h,
    w, d)$ and produces an output of shape $(w', d')$.
    A summarizer produces an output of shape $(|S|)$.
    Encoders and summarizers are
    typically modeled by neural networks and are trained end-to-end.}
  \label{fig:encoder-summarizer}
\end{figure}

Line-level script identification takes a line image as input and produces a
script code. We model the relationship with a probability $P(s|\bvec{x})$. Since
$\bvec{x}$ is inherently a sequence, we solve the problem as a sequence-to-label
problem with two components that are trained end-to-end: \emph{Encoder} and
\emph{Summarizer}. Encoders are used to convert a line image into a sequence of
features and summarizers are used to aggregate the sequence to perform the
classification. Fig~\ref{fig:encoder-summarizer} visualizes the concept. An
encoder $E(\bvec{x})$ takes $\bvec{x}$ as input and outputs a sequence of
features $\bvec{h}$ that is a tensor of shape $(w', d')$ where $w'$ is the length and $d'$
is the dimensionality of the features. A summarizer $F(\bvec{h}, s)$ transforms
the feature sequence into a set of scalar values, one for each script
$s$. These values are then transformed via \emph{softmax} into the posterior
distribution $P(s|\bvec{x})$ with $E(\bvec{x})$ and
$F(\bvec{h}, s)$:
\begin{align}
  \label{eq:encoder-summarizer}
  P(s|\bvec{x}) &= \frac{\exp(F(E(\bvec{x}), s))}{\sum_{s'}\exp(F(E(\bvec{x}),
    s'))}.
\end{align}
We train all model variations using cross-entropy loss.
The following sections describe the encoder and
summarizers considered in this work.

\subsection{Encoder}
\label{sec:encoder}

We use the inception-style convolutional neural network shown in
Fig.~\ref{fig:convolutional-layer} as the encoder for all summarizers.
It is identical to the encoder layer of the OCR network described in
Section~\ref{sec:text-line-recognition-with-scriptid}.
This allows a fair comparison between all summarizer variants.
Note that further optimization is possible: we can reduce the number of
parameters (e.g. fewer channels in every convolutional layer) without
hurting accuracy since classification to 30 scripts needs fewer
parameters than accurate character recognition for a line of symbols.

\subsection{Summarizer}
\label{sec:summarizer}

\subsubsection{Max}

The Max summarizer produces a tensor of shape $(|S|)$ for each index
of a sequence $\bvec{h}$ and takes the maximum  per script over the
sequence which has $w'$ values for each script.
We represent the function to compute the
tensor of shape $(w', |S|)$ from $\bvec{h}$ as $L(\bvec{h})$. We model
$L(\bvec{h})$ with a neural network and use the topology shown in
Fig.~\ref{fig:logits-net}. Formally, we define the summarizer as
follows:
\begin{align}
  \label{eq:summarizer-max}
  F^{\text{max}}(\bvec{h}, s) &\defeq \max_{i}{l_{i,s}},
\end{align}
where $\bvec{l}_{*,s} = (l_{1, s}, l_{2, s}, \dots, l_{w', s}) = L(\bvec{h})$.

\subsubsection{Mean}

The Mean summarizer computes the mean of $L(\bvec{h})$ for each script as
follows:
\begin{align}
  \label{eq:summarizer-mean}
  F^{\text{mean}}(\bvec{h}, s) &\defeq \frac{1}{|l_{*,s}|}\sum_{i}{l_{i,s}}.
\end{align}
Max and Mean summarizers correspond to the methods proposed
in~\cite{BaoguangShi:2015:ICDAR}.

\subsubsection{Gate}

\begin{figure}[!t]
\centering
\subfloat[$L(\bvec{h})$]{\includegraphics[width=3.5cm]{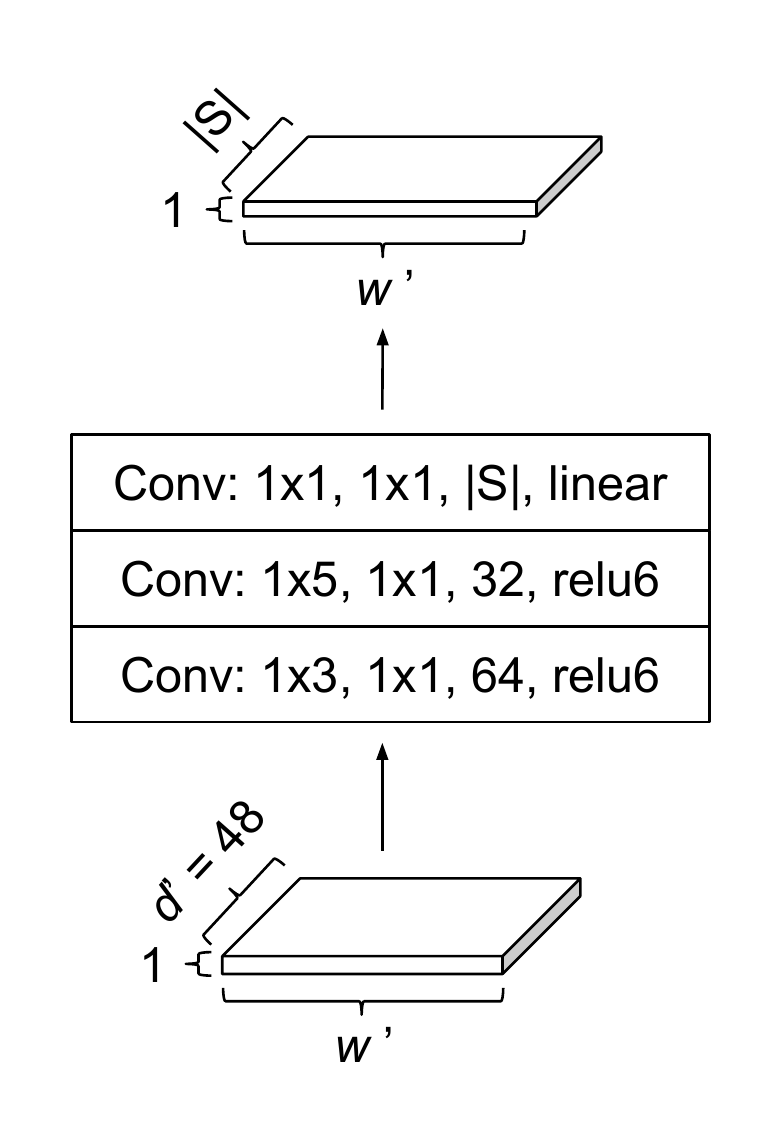}
\label{fig:logits-net}}
\hfil
\subfloat[$G(\bvec{h})$]{\includegraphics[width=3.5cm]{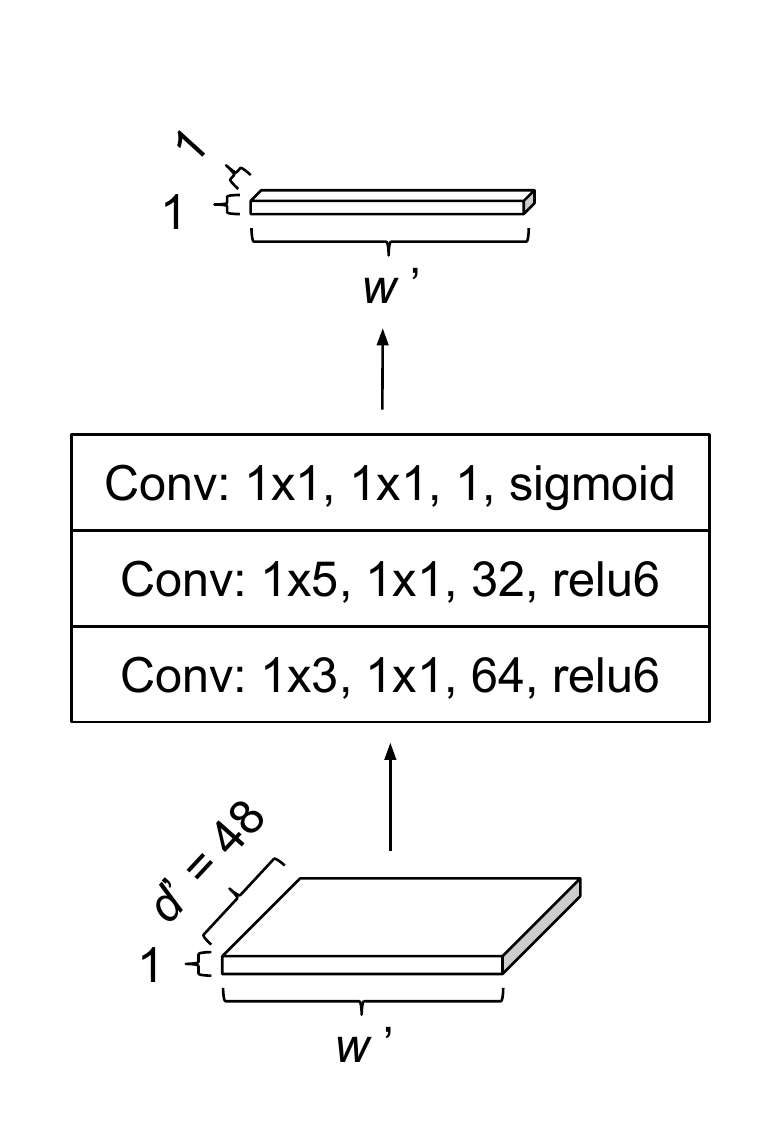}
\label{fig:gate-net}}
\caption{Topology of $L(\bvec{h})$ and $G(\bvec{h})$. $\text{linear}(x) =
  x$. $\text{sigmoid}(\bvec{x}) = 1 / (1 + \exp(-\bvec{x}))$. The SAME padding
  is used for all operations. See Fig.~\ref{fig:convolutional-layer} for the
  other notations.}
\end{figure}

The Gate summarizer computes the mean of $L(\bvec{h})$ for each script as the
Mean summarizer does. However, it also utilizes an attention-like gating
mechanism to select indices which should be considered in the final output. We
denote a function as $G(\bvec{h})$ to convert a tensor of shape $(w', d')$ to a
tensor of shape $(w')$. We model $G(\bvec{h})$ with a neural network and use the
topology shown in Fig.~\ref{fig:gate-net}. Formally, we define the summarizer as
follows:
\begin{align}
  \label{eq:summarizer-gate}
  F^{\text{Gate}}(\bvec{h}, s) &\defeq \frac{1}{\sum_i g_i}\sum_{i}{g_i \cdot l_{i,s}},
\end{align}
where $\bvec{g} = (g_{1}, g_{2}, \dots, g_{w'}) = G(\bvec{h})$.

\subsubsection{LSTM}

\begin{figure}[!t]
  \centering
  \includegraphics[width=3.5cm]{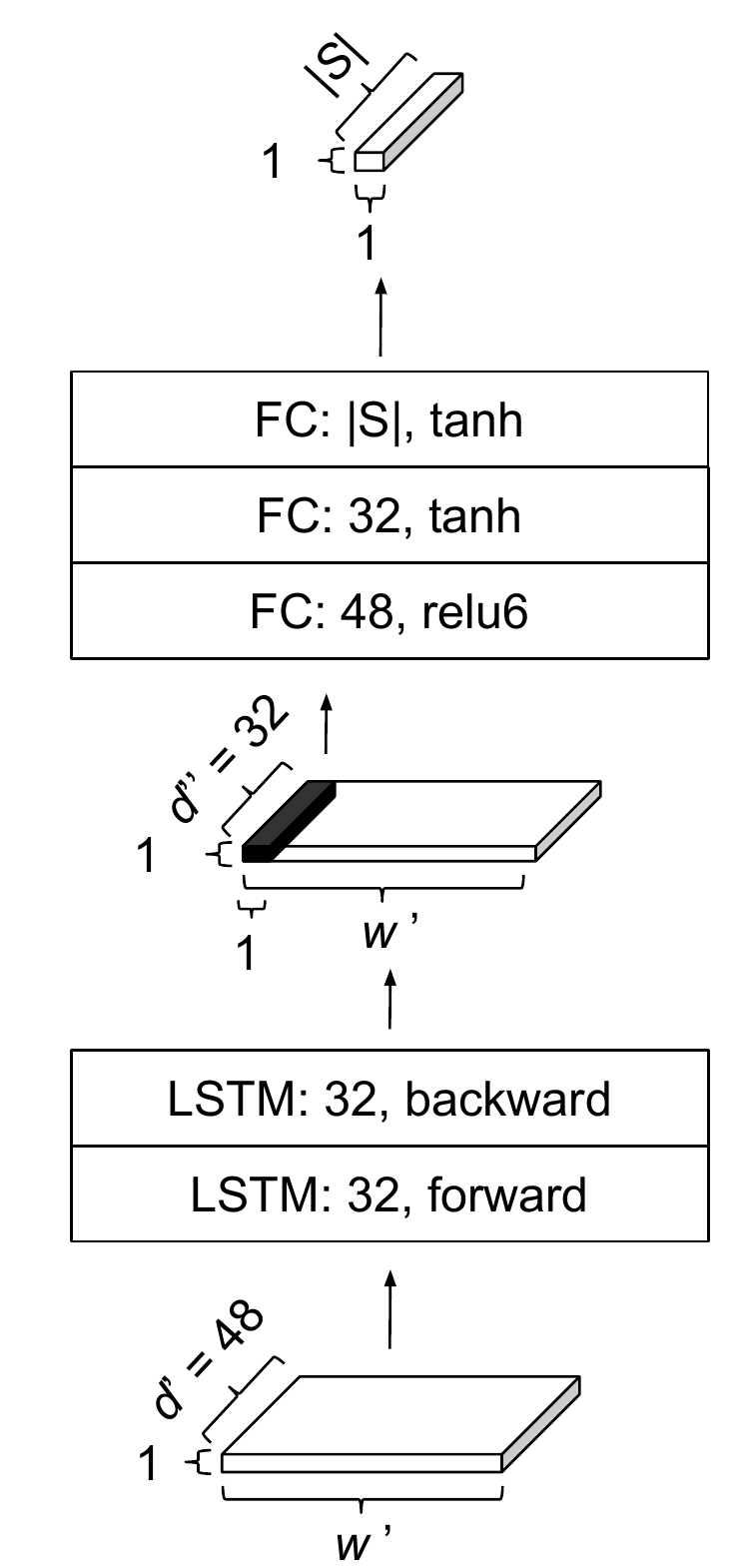}
  \caption{Topology of $M(\bvec{h})$. ``LSTM: $n, d$'' represents
    an LSTM layer
    with $n$ nodes in the direction $d$.
    ``FC: $n, f$'' represents a
    fully-connected layer with $n$ nodes and an activation
    function $f$. The
    values at the first index of the output of the backward LSTM
    are used as the
    aggreated values (depicted in black).}
  \label{fig:lstm-net}
\end{figure}

In \cite{IlyaSutskever:2014:NIPS}, long short-term memory
(LSTM)~\cite{Hochreiter:1997:LSM:1246443.1246450} is used to encode a sequence
of inputs to a single output.
Similar techniques can be used to formulate a
summarizer. The LSTM summarizer uses a forward LSTM followed
by a backward LSTM
to aggregate the sequence.
Both forward and backward LSTMs consume a sequence
and produce another sequence.
We use the values at the first index of the output
of the backward LSTM as the aggregated values.
We further apply a sequence of
fully-connected layers to obtain the final output of shape $(|S|)$.
We represent these operations as $M(\bvec{h})$.
The topology of $M(\bvec{h})$ is shown in Fig.~\ref{fig:lstm-net}.
Formally, we define the summarizer as follows:
\begin{align}
  \label{eq:summarizer-lstm}
  F^{\text{LSTM}}(\bvec{h}, s) &\defeq l_{s}^{\text{LSTM}},
\end{align}
where $\bvec{l}^{\text{LSTM}} = M(\bvec{h})$.

\subsection{Baseline}
\label{sec:ctc-summarizer}

The OCR system described in
Section~\ref{sec:text-line-recognition-with-scriptid} is repurposed for script
identification as proposed in~\cite{DmitriyGenzel:2013:HSI:2505377.2505382} by
training it to output script codes instead of characters (1 per character).
We implemented exactly the heuristic described
in~\cite{DmitriyGenzel:2013:HSI:2505377.2505382} to determine the \emph{dominant
  script} of a line from the sequence of script codes.  This heuristic gives
each character a vote. Majority decides a unique script label per line.
Some characters such as spaces and digits are ignored. Characters common in
Japanese, Chinese Simplified and Chinese Traditional count towards all 3
scripts. Full details can be found
in~\cite{DmitriyGenzel:2013:HSI:2505377.2505382}. We refer to this baseline
method as \emph{Base} in the following sections.

\subsection{Model Sizes}
\label{sec:model-sizes}

\begin{table}[t]
  \centering
\begin{tabular}{|r r|r|}
\hline
& & Weights\\
\hline
\multicolumn{2}{|c|}{Encoder} & 2,905,152\\
\hline
\multirow{4}{*}{Summarizers} & Gate & 39,968\\
& Lstm & 23,103\\
& Sum & 20,479\\
& Max & 20,479\\
\hline\end{tabular}

  \caption{Model Sizes}
  \label{tab:model-sizes}
\end{table}

Table~\ref{tab:model-sizes} shows model sizes. All models have close to
3 million weights. Most are used by the convolutional layers
in the Encoder. The same Encoder is used between all end-to-end models.
It is identical to feature-producing network (prior to CTC) in
the repurposed OCR model that provides per-character
script codes in the Baseline system. Summarizers are small compared to
overall model size. They account for around 1\% of total parameter size.

\section{Evaluation Data and Method}

\begin{figure}
\centering
\includegraphics[width=8cm]{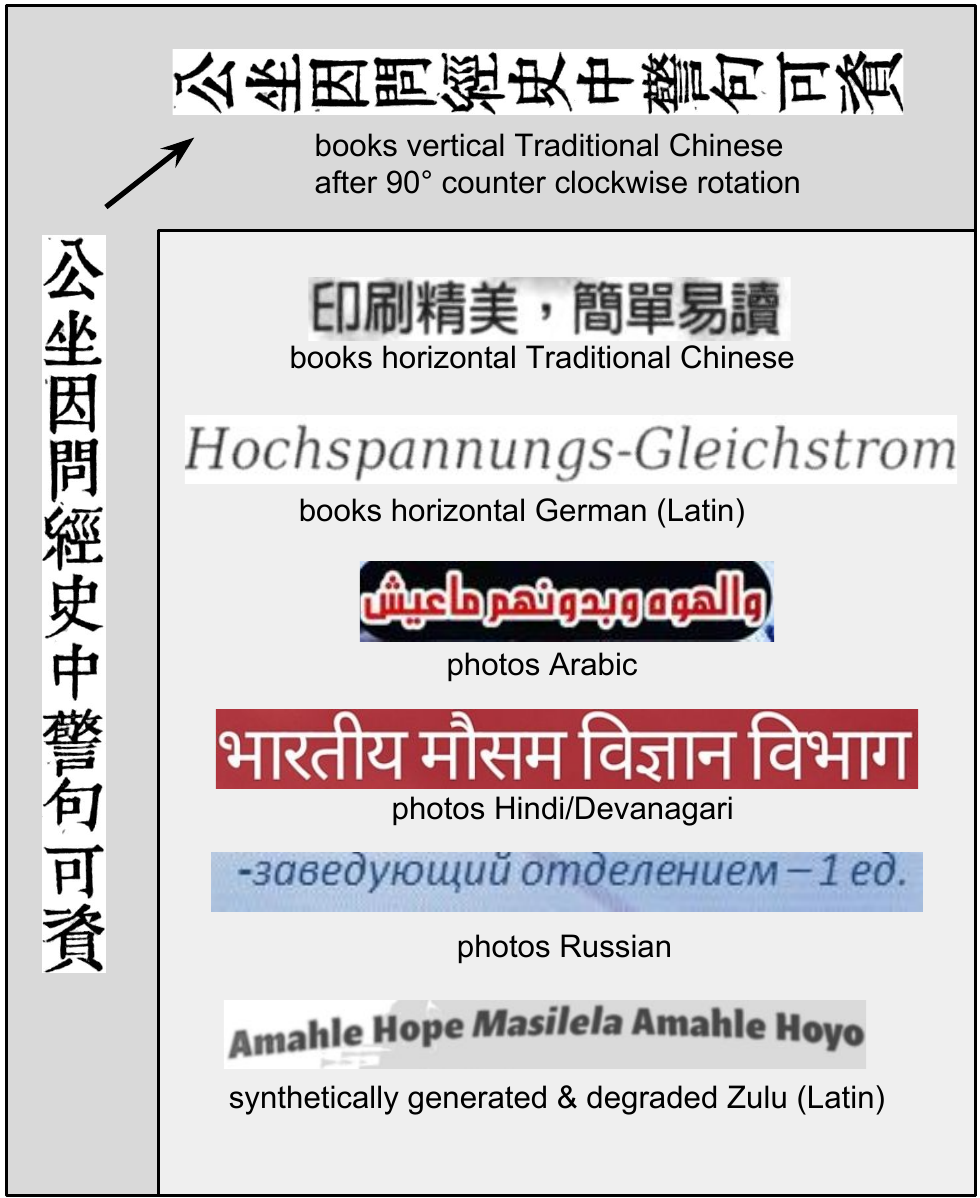}
\caption{Examples of text line images from different domains.}
\label{fig:text-line-images}
\end{figure}

Fig.~\ref{fig:text-line-images} shows examples of text line images
from 3 domains. Books-horizontal contains dense, book-like
lines from 30 scripts. Books-vertical has book-like vertical
lines in Korean, Japanese, and Chinese Traditional. While other
languages also have vertical text, those characters
are usually oriented perpendicular to the line. The line
becomes horizontal after 90 degree rotation. Hence we only need
separate script models for truly vertical script with
characters oriented along the line. The photos domain has
text lines from photo-like images (e.g., born-digital, natural-scene). It is
typically more challenging than book-like images.

\begin{table}[t]
  \centering
\begin{tabular}{|r|r|r|r|r|}
\hline
 &  & & \multicolumn{2}{c|}{number of lines}\\
domain & scripts & langs & eval & train\\ 
\hline
\multirow{5}{*}{photos} & Latin & 1 & 9,235 & 192,383\\ 
 & Cyrillic & 1 & 3,981 & 0\\ 
 & Japanese & 1 & 2,863 & 0\\ 
 & Arabic & 1 & 2,442 & 0\\ 
 & Devanagari & 1 & 2,166 & 0\\ 
\hline
Sum & 5 & 5 & 20,687 & 192,383\\ 
\hline
\hline
\multirow{3}{*}{books vertical} & Chinese Trad. & 1 & 1,790 & 39,779\\ 
 & Japanese & 1 & 1,767 & 51,164\\ 
 & Korean & 1 & 921 & 37,644\\ 
\hline
Sum & 3 & 3 & 4,478 & 128,587\\ 
\hline
\hline
\multirow{30}{*}{books horizontal} & Latin & 163 & 61,883 & 62,649\\ 
 & Cyrillic & 28 & 24,088 & 92,530\\ 
 & Devanagari & 4 & 11,649 & 40,411\\ 
 & Arabic & 6 & 10,900 & 40,337\\ 
 & Tibetan & 2 & 8,860 & 15,125\\ 
 & Syriac & 1 & 3,370 & 14,893\\ 
 & Hebrew & 2 & 3,184 & 15,107\\ 
 & Ethiopean & 2 & 3,122 & 14,737\\ 
 & Georgian & 1 & 3,002 & 14,917\\ 
 & Cherokee & 1 & 2,935 & 14,610\\ 
 & Bengali & 2 & 2,809 & 14,819\\ 
 & Myanmar & 1 & 2,781 & 14,890\\ 
 & Sinhala & 1 & 2,493 & 15,005\\ 
 & Oriya & 1 & 1,626 & 15,226\\ 
 & Thaana & 1 & 1,363 & 14,526\\ 
 & Laoo & 1 & 1,251 & 15,409\\ 
 & Gujarati & 1 & 1,184 & 14,807\\ 
 & Gurmukhi & 1 & 1,166 & 14,516\\ 
 & Telugu & 1 & 1,152 & 15,064\\ 
 & Kannada & 1 & 1,150 & 14,826\\ 
 & Khmer & 1 & 1,133 & 14,926\\ 
 & Malayalam & 1 & 1,124 & 14,795\\ 
 & Thai & 1 & 1,062 & 14,976\\ 
 & Armenian & 1 & 1,048 & 15,029\\ 
 & Tamil & 1 & 989 & 15,510\\ 
 & Greek & 2 & 932 & 15,016\\ 
 & Korean & 1 & 898 & 56,973\\ 
 & Chinese Simpl. & 1 & 834  & 63,474\\ 
 & Japanese & 1 & 802 & 62,134\\ 
 & Chinese Trad. & 1 & 796 & 57,429\\ 
\hline
Sum & 30 & 232 & 159,586 & 804,666\\ 
\hline
\hline
Overall Sum & 30 & 232 & 183,388 & 1,125,636\\ 
\hline\end{tabular}

  \caption{Data Description}
  \label{tab:data-description}
\end{table}

Table~\ref{tab:data-description} breaks down evaluation and training
data. Non Latin scripts have fully synthetic training data.
We digitally render text from Wikipedia and other web sources with
added noise and distortion. For example, see the Zulu line in
Fig.~\ref{fig:text-line-images}. Latin has in addition real, photo-like data
from advertisements, for about 75\% of total training data.
Evaluation data is mostly
non-synthetic. Model definition and training use
TensorFlow~\cite{AbadiMartin:2016:OSDI}.

Evaluation data has labeled real text line images from
232 languages in 30 scripts. Images come from web pages,
smartphone photos, and book scans.
Each labeled line image has transcribed text and bcp47 language code,
which determines script label. 82K out of 183K lines in
the horizontal books test set are synthetic.
The other 101K are images of books or photos
labeled by human raters. The 20K photo-like lines
in 5 languages/scripts are all human rated.
70\% of supported languages use Latin script,
but only 39\% of evaluation text lines are Latin. Some scripts
require more data, e.g. vertical Japanese or Arabic.

On script labels for multi-script lines:
we first obtained a nominal language label by collecting
test data in \emph{a priori known} language corpuses.
We then send text lines to
language-capable human raters for verification of the language and
script label. The raters also perform transcription of the text line.
In some cases a line that is identified and confirmed to be in a
particular language may contain a long phrase from a different script.
To mitigate this issue we apply a filtering step in preparing our
evaluation data. For example, we drop a line in the Japanese
script eval set with less than 50\% Japanese symbols
(e.g. too much Latin text).

\section{Results}
\label{sec:results}

\begin{table}[t!]
  \centering
\begin{tabular}{|r|p{2.2cm}|p{2.2cm}|}
\hline
 & \multicolumn{2}{|c|}{Confusions over 1\%} \\ 
 & Base & Gate\\ 
\hline
Latin & - & -\\ 
\hline
Cyrillic & Latin:4.2\% & Latin:1.8\%\\ 
\hline
Devanagari & - & -\\ 
\hline
Arabic & - & -\\ 
\hline
Tibetan & Latin:1.3\% & -\\ 
\hline
Japanese & Chinese-Sim:1.9\% Chinese-Tr:2.6\% & Chinese-Tr:3.9\%\\ 
\hline
Syriac & Latin:2.9\% Arabic:1.2\% & Latin:1.2\% NonText:1.1\%\\ 
\hline
Ethiopean & - & -\\ 
\hline
Hebrew & - & -\\ 
\hline
Georgian & - & -\\ 
\hline
Cherokee & Latin:5.2\% & Latin:2.6\%\\ 
\hline
Myanmar & - & -\\ 
\hline
Bengali & Latin:2.8\% & Latin:1.6\%\\ 
\hline
Chinese-Tr & Chinese-Sim:7.3\% & -\\ 
\hline
Sinhala & - & -\\ 
\hline
Korean & Chinese-Tr:1.3\% & Chinese-Tr:1.2\%\\ 
\hline
Oriya & Latin:4.4\% & Latin:2.2\%\\ 
\hline
Thaana & - & -\\ 
\hline
Laoo & Latin:3.5\% & Latin:1.6\% Telugu:1.4\%\\ 
\hline
Gujarati & - & -\\ 
\hline
Gurmukhi & Arabic:2.4\% Latin:1.8\% & Arabic:2.5\% Latin:1.1\%\\ 
\hline
Telugu & - & -\\ 
\hline
Kannada & - & -\\ 
\hline
Khmer & Latin:15.7\% Thai:2.6\% & Latin:5.5\% Laoo:1.7\%\\ 
\hline
Malayalam & Latin:1.2\% & -\\ 
\hline
Greek & Latin:12.3\% & Latin:1.4\%\\ 
\hline
Thai & - & -\\ 
\hline
Armenian & Latin:21.1\% & Latin:2.3\%\\ 
\hline
Tamil & Latin:3.4\% & -\\ 
\hline
Chinese-Sim & Chinese-Tr:1.7\% & Chinese-Tr:3.7\%\\ 
\hline\end{tabular}

  \caption{Script confusions over 1\% on books data }
  \label{tab:confusion}
\end{table}

Table~\ref{tab:results} shows results for the 5 Script identification
summarization techniques described in section ~\ref{sec:summarizer}.
We give the script identification error rate per domain. This is
defined as number of lines in which script identification
disagrees with the script label, divided by total number of
lines. We also show OCR error rate due to script identification.
This is end-to-end character error rate for the script identification
technique followed
by our best per-script OCR model, subtracted by ``Oracle-script''
OCR error rate. This subtraction isolates OCR errors from script
misidentification errors.
The Oracle-script system simulates ``perfect'' script identification
by using the script label.
We show results for the photo domain, vertical book-like data for
Korean/Japanese/Chinese Traditional, and horizontal book-like data.

Overall, the Gate system is the clear winner. Compared to Base
it reduces script identification error rate by 16\% relative (from
3.7\% to 3.1\%).  The impact on end-to-end character recognition error
from script misidentification goes down 33\%, from 2.1\% to 1.4\%.

On subdomains, Gate beats other summarization methods with
one exception. Max is slightly better on books-vertical, in
particular on Korean. However, this does not result in significant
character error rate reduction (0.1\% on books-vertical). Overall,
Max does not perform well. It is even
slightly worse than Base on overall script identification error rate.

No technique beats all others on every script. Every technique achieves lowest
script identification error on at least one script. Base wins on Latin, LSTM on
Bengali, Mean on Cherokee, Max on Korean, and Gate on Khmer, among
others. In aggregate all non-Base techniques excluding Max
beat Base. Gate is the best among them.

Table~\ref{tab:confusion} has the top script
confusions\footnote{Full confusion matrices in http://arxiv.org/abs/1708.04671.}
in books data per script for Base and Gate.
We list all scripts accounting for more
than 1\% errors in any script. Mostly these are unsurprising. For
instance Cyrillic has many characters that look like similar Latin
characters. Japanese as well as Chinese Simplified retain the use
of certain Chinese Traditional characters. Therefore these misidentifications
are expected. Similarly, many non-Latin texts use Latin
characters for names instead of transliterating.
This explains Latin confusion.
Gate gets rid of confusion or greatly diminishes it for all scripts
except Chinese Simplified. The confusion with Chinese
Traditional increases, congruent with the increased
script identification error rate in Table~\ref{tab:results}.

Base is best on Japanese and second best on Korean for the books-vertical
test set.  However, on Chinese Traditional it performs so poorly that the
mean script identification error on the domain is twice that of Gate and Max.

\begin{table*}
  \centering
\begin{tabular}{|r|r|r r r r r ||r r r r r||r|}
\hline
& & \multicolumn{5}{|c||}{Script ID Error Rate \%} & \multicolumn{5}{|c||}{OCR Error \% due to Script ID} & \multicolumn{1}{|c|}{Oracle-script}\\
domain & script & Base & Gate & LSTM & Mean & Max & Base & Gate & LSTM & Mean & Max & OCR Err\%\\ 
\hline
\multirow{6}{*}{photos} & Latn & 1.0 & 1.2 & \textbf{0.8} & \textbf{0.8} & 1.4 & \textbf{0.8} & \textbf{0.8} & \textbf{0.8} & \textbf{0.8} & 0.9 & 10.5\\ 
 & Cyrl & 49.5 & \textbf{28.7} & 33.5 & 35.1 & 30.8 & 28.0 & \textbf{11.0} & 13.0 & 14.5 & 12.4 & 11.4\\ 
 & Jpan & \textbf{25.5} & 30.1 & 39.7 & 31.2 & 30.4 & \textbf{8.5} & 12.3 & 15.5 & 13.5 & 11.2 & 21.1\\ 
 & Arab & \textbf{19.0} & 23.5 & 26.7 & 26.0 & 27.7 & \textbf{6.8} & 7.0 & 7.4 & 7.9 & 9.0 & 15.0\\ 
 & Deva & \textbf{17.6} & 22.9 & 21.5 & 27.9 & 23.7 & \textbf{4.5} & 6.2 & 5.5 & 7.8 & 7.2 & 16.3\\ 
\cline{2-13}
 & Mean & 17.6 & \textbf{15.4} & 17.7 & 17.5 & 16.5 & 8.2 & \textbf{5.6} & 6.4 & 6.7 & 6.1 & 13.3\\ 
\hline
\multirow{4}{*}{books vertical} & Hant & 8.9 & \textbf{0.3} & 1.6 & 1.3 & 1.1 & 1.5 & \textbf{0.0} & 0.2 & 0.1 & 0.1 & 10.5\\ 
 & Jpan & \textbf{3.1} & 3.4 & 3.6 & 4.0 & 3.2 & \textbf{0.2} & \textbf{0.2} & \textbf{0.2} & 0.3 & 0.4 & 6.1\\ 
 & Kore & 2.9 & 4.1 & 4.8 & 5.2 & \textbf{1.4} & 1.0 & 1.7 & 2.1 & 2.0 & \textbf{0.5} & 7.3\\ 
\cline{2-13}
 & Mean & 5.4 & 2.3 & 3.0 & 3.2 & \textbf{2.0} & 0.9 & 0.4 & 0.6 & 0.6 & \textbf{0.3} & 8.1\\ 
\hline
\multirow{31}{*}{books horizontal} & Latn & \textbf{0.4} & 1.1 & 1.1 & 0.9 & 1.6 & \textbf{0.9} & 1.0 & 1.1 & 1.0 & 1.2 & 4.1\\ 
 & Cyrl & 3.6 & \textbf{1.3} & 2.2 & 2.4 & 1.9 & 2.4 & \textbf{0.7} & 1.2 & 1.5 & 1.3 & 2.2\\ 
 & Deva & \textbf{0.9} & 1.0 & 1.0 & 1.2 & 1.1 & \textbf{0.2} & \textbf{0.2} & 0.3 & 0.4 & 0.3 & 5.7\\ 
 & Arab & \textbf{0.9} & 1.1 & 1.7 & 1.6 & 1.8 & \textbf{0.2} & 0.3 & 0.4 & 0.4 & 0.6 & 6.6\\ 
 & Tibt & \textbf{2.4} & 2.9 & 3.2 & \textbf{2.4} & 3.1 & 1.4 & 1.7 & 1.6 & \textbf{1.2} & 1.9 & 4.7\\ 
 & Syrc & \textbf{4.4} & 4.6 & 4.5 & 4.7 & 6.0 & \textbf{1.2} & 1.4 & 1.3 & 1.3 & 2.0 & 1.9\\ 
 & Hebr & 0.2 & \textbf{0.1} & 0.3 & 0.4 & \textbf{0.1} & \textbf{0.1} & \textbf{0.1} & 0.2 & 0.2 & \textbf{0.1} & 5.2\\ 
 & Ethi & \textbf{1.0} & 1.3 & 2.9 & 2.2 & 2.4 & \textbf{0.5} & 0.7 & 1.2 & 0.9 & 1.1 & 0.5\\ 
 & Geor & \textbf{0.7} & 1.0 & 1.6 & 1.1 & 1.7 & \textbf{0.4} & 0.7 & 1.1 & 0.7 & 1.2 & 0.4\\ 
 & Cher & 5.2 & 3.5 & 4.3 & \textbf{3.1} & 5.0 & 3.9 & 2.6 & 3.4 & \textbf{2.2} & 3.5 & 0.4\\ 
 & Beng & 3.2 & 3.1 & \textbf{3.0} & 3.3 & 3.1 & 0.9 & 0.8 & \textbf{0.7} & 0.9 & 1.0 & 5.0\\ 
 & Mymr & \textbf{1.5} & 1.7 & 1.8 & 1.6 & 2.1 & \textbf{0.4} & 0.7 & 0.7 & 0.6 & 0.8 & 4.1\\ 
 & Sinh & \textbf{1.2} & 1.6 & \textbf{1.2} & 1.4 & 1.5 & 0.3 & 0.5 & \textbf{0.2} & 0.3 & 0.4 & 9.2\\ 
 & Orya & 5.0 & 4.0 & \textbf{2.8} & 4.1 & 5.7 & 2.7 & 1.5 & \textbf{1.3} & 1.9 & 2.3 & 6.0\\ 
 & Thaa & 0.7 & \textbf{0.6} & 1.4 & 0.8 & 0.7 & 0.3 & \textbf{0.1} & 0.4 & \textbf{0.1} & 0.2 & 1.1\\ 
 & Laoo & 6.7 & 4.6 & 8.9 & \textbf{4.1} & 9.6 & 2.1 & 1.2 & 3.5 & \textbf{1.1} & 4.5 & 14.6\\ 
 & Gujr & 1.0 & \textbf{0.4} & 0.6 & 0.8 & 0.5 & 0.5 & \textbf{0.3} & \textbf{0.3} & 0.4 & \textbf{0.3} & 7.4\\ 
 & Guru & 4.7 & 4.3 & 4.3 & \textbf{4.1} & 4.3 & 2.4 & 2.3 & 2.3 & 2.3 & \textbf{2.0} & 7.9\\ 
 & Telu & 1.0 & 0.4 & 0.6 & 1.0 & \textbf{0.3} & 0.4 & \textbf{0.2} & \textbf{0.2} & \textbf{0.2} & \textbf{0.2} & 7.7\\ 
 & Knda & 1.0 & 0.7 & \textbf{0.3} & 0.5 & 0.9 & 0.4 & 0.3 & \textbf{0.0} & 0.1 & 0.3 & 12.8\\ 
 & Khmr & 18.6 & \textbf{7.9} & 11.2 & 15.3 & 20.6 & 7.3 & \textbf{1.8} & 2.6 & 5.4 & 8.3 & 31.6\\ 
 & Mlym & 1.2 & \textbf{0.1} & 0.2 & 0.5 & 0.6 & 1.0 & \textbf{0.0} & 0.1 & 0.2 & 0.5 & 5.7\\ 
 & Thai & 0.3 & 0.4 & \textbf{0.1} & 0.3 & 0.8 & 0.2 & 0.3 & \textbf{0.1} & 0.4 & 0.6 & 3.1\\ 
 & Armn & 21.2 & \textbf{2.6} & 3.7 & 12.2 & 6.1 & 15.8 & \textbf{1.3} & 1.9 & 8.5 & 3.4 & 4.2\\ 
 & Taml & 3.4 & \textbf{0.4} & 0.5 & 1.4 & 1.0 & 2.2 & \textbf{0.2} & \textbf{0.2} & 0.6 & 0.7 & 5.8\\ 
 & Grek & 12.4 & 1.4 & 1.6 & \textbf{1.2} & 2.9 & 9.1 & 1.0 & 1.1 & \textbf{0.7} & 2.0 & 4.8\\ 
 & Kore & 1.3 & 0.3 & 0.6 & 0.7 & \textbf{0.1} & 0.5 & 0.1 & 0.1 & 0.2 & \textbf{0.0} & 2.5\\ 
 & Hans & \textbf{1.8} & 5.3 & 6.2 & 2.9 & 4.8 & \textbf{0.1} & 0.5 & 1.0 & 0.2 & 0.6 & 2.4\\ 
 & Jpan & 8.2 & 7.2 & 7.5 & 10.1 & \textbf{7.0} & 0.8 & \textbf{0.6} & \textbf{0.6} & 1.1 & \textbf{0.6} & 4.9\\ 
 & Hant & 6.0 & \textbf{2.1} & 2.5 & 5.4 & 4.4 & 1.1 & \textbf{0.2} & 0.3 & 0.7 & 0.7 & 2.8\\ 
\cline{2-13}
 & Mean & 1.9 & \textbf{1.5} & 1.8 & 1.8 & 2.2 & 1.3 & \textbf{0.9} & 1.0 & 1.0 & 1.2 & 4.4\\ 
\hline
\multicolumn{2}{|c|}{Overall line-weighted mean} & 3.7 & \textbf{3.1} & 3.6 & 3.5 & 3.8 & 2.1 & \textbf{1.4} & 1.6 & 1.7 & 1.8 & 5.5\\ 
\hline\end{tabular}

  \caption{Results}
  \label{tab:results}
\end{table*}

Photo data is inherently more challenging than books data
which tends to be cleaner and less
distorted.  When the script is
given, character recognition error rate is 2.5 to 5 times higher in
photos than in books, as shown in the Oracle column. For example,
Latin has 10.5\% error rate in photos, 4.1\% in
books. Cyrillic has the largest difference between photos (11.4\%) and
books (2.2\%). We plan to get more photo training data to reduce this gap.

Script identification error rates for Latin are remarkably lower than the
other 4 scripts in the photos category. We believe this is primarily
caused by the presence of real photo-like Latin training data.
When we remove this data, the script identification error rate
for Gate summarizer on the Latin data set increases to 7\%.
We expect that adding photo-like script
identification training data for other scripts will
lower error rates further.

Photos result in aggregate agree with overall
results.
Compared to Base, Gate reduces script identification error rate by 12.5\%
relative (from 17.6\% to 15.4\%).  The impact on end-to-end character
recognition error is a 32\% relative reduction, from 8.2\% to
5.6\%.

Cyrillic is an outlier. It has 11.0\% end-to-end error rate
with best script identification on top of 11.4\% OCR error
rate for Oracle-script.
Base error is high on Cyrillic photos.  The 49.5\% script identification
error rate is caused by Latin-looking characters having equal weight
as distinctive Cyrillic characters in the voting heuristic of Base.
Gate is better able to ignore confusable characters (more precisely,
non-script-informative feature elements) through its attention mechanism.

\section{Discussion}
\label{sec:discussion}

\begin{figure*}[!t]
\centering
\subfloat[Gate]{\includegraphics[width=5.8cm]{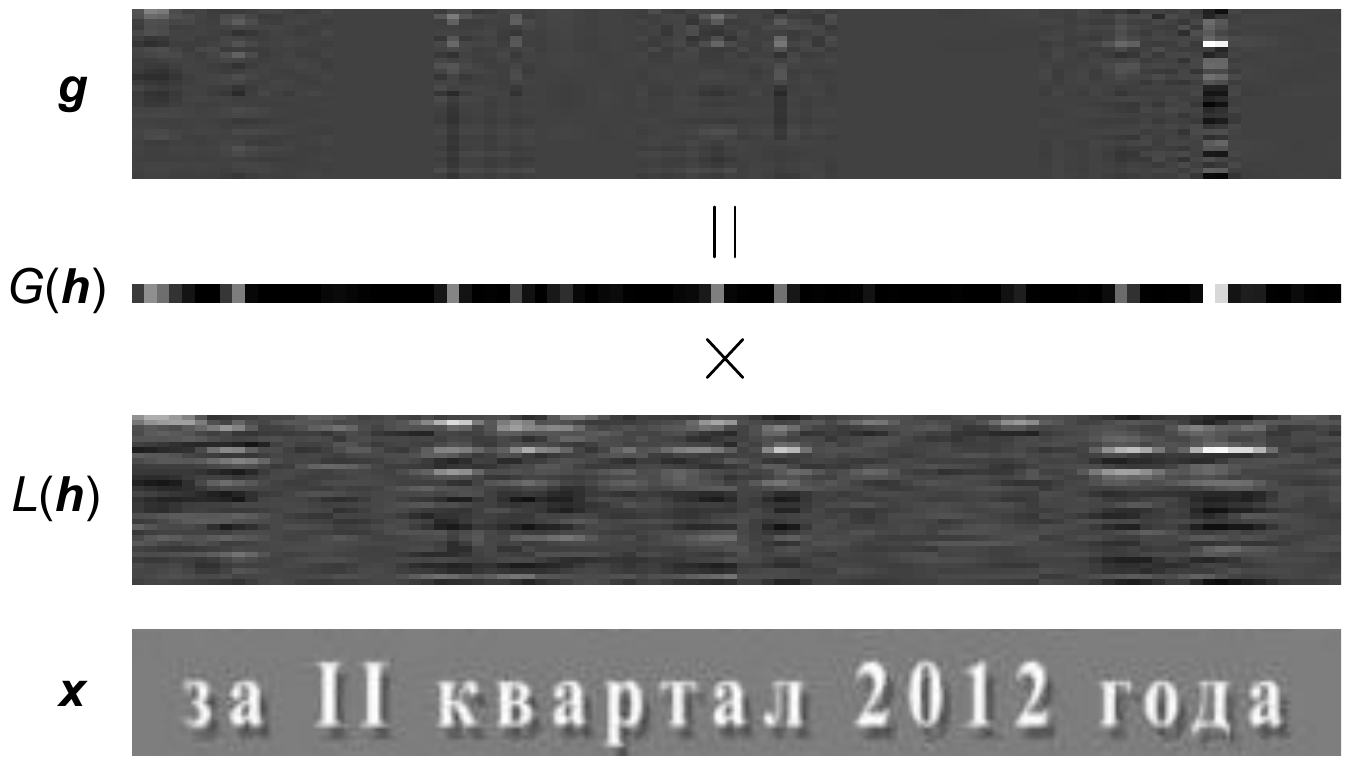}
\label{fig:logits-gate}}
\hfil
\subfloat[Mean]{\includegraphics[width=5.8cm]{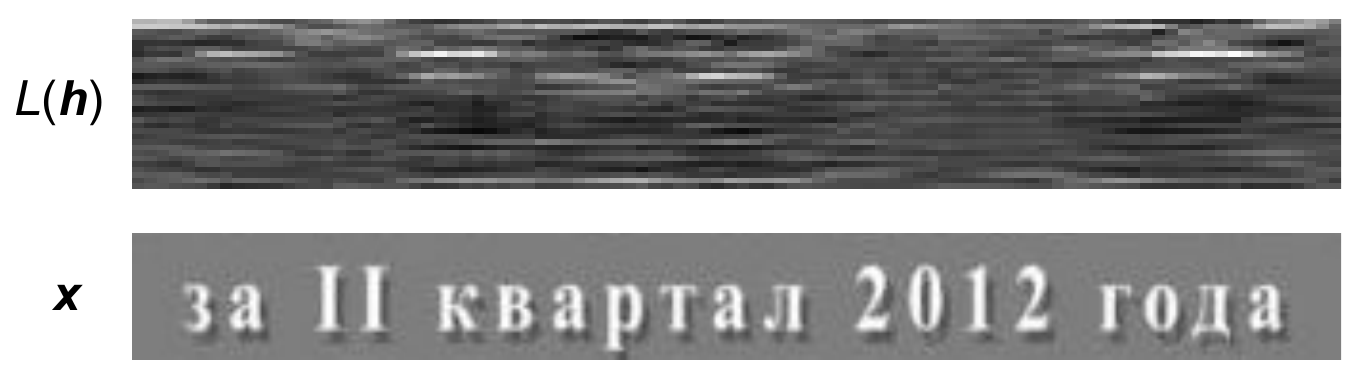}
\label{fig:logtis-mean}}
\hfil
\subfloat[Max]{\includegraphics[width=5.8cm]{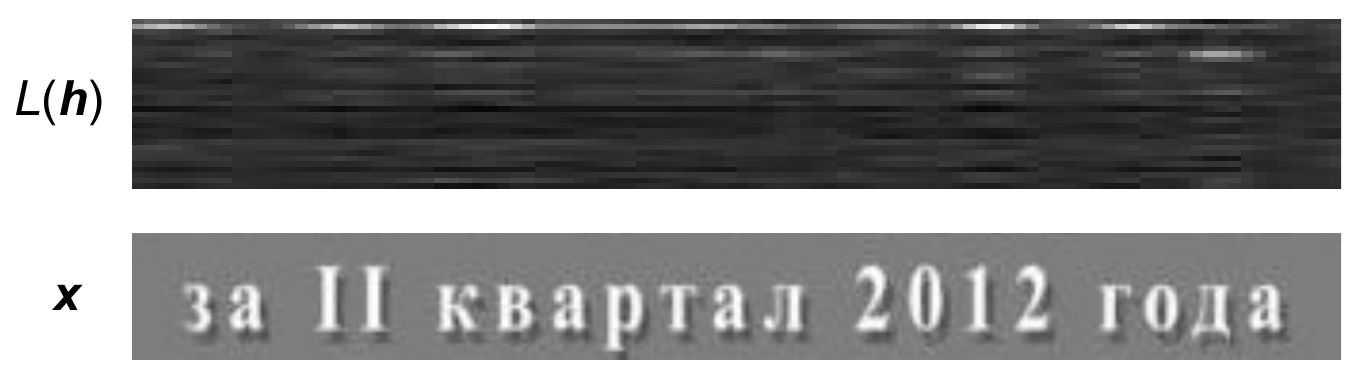}
\label{fig:logits-max}}
\caption{Visualization of the activation maps of Gate, Mean and Max
  summarizers. Brigher pixels mean greater values. Each row corresponds to a
  script. The input image is in Cyrillic script and obtained from the photo
  dataset. For this particular example, Gate and Max correctly predict the
  script while Mean mispredicts it as Latin. The gating mechanism allows for
  Gate to attend to Cyrillic characters (e.g. the second to last character) and
  ignore irrelevant characters such as digits. }
\label{fig:logits-visualization}
\end{figure*}

Fig.~\ref{fig:logits-visualization} visualizes the activation maps of Gate, Mean
and Max summarizers. The activation map of $L(\bvec{h})$ looks
different for each summarizer. This indicates that each summarizer exerts
distinct requirements on $L(\bvec{h})$. The Mean activation map is fuzzier than
for other summarizers. This makes sense considering that the Mean summarizer
essentially performs voting. Even small values contribute to the final
decision. In contrast, the Max summarizer produces the sharpest activation
map. This makes sense since the Max summarizer considers only the
strongest signal for each script. The Gate summarizer lies between the two in
terms of the sharpness of the activation map. The gating
mechanism (top rectangle $\bvec{g}$) suppresses regions where the characters are
not Cyrillic. It considers only relevant regions of the input.  For instance,
the activation map of $L(\bvec{h})$ has activity on digits and the
Latin-like character "a". However, the activation map of $\bvec{g}$
suppresses this
activity and highlights distinctive Cyrillic characters.  This illustrates
why the Gate summarizer performed so well.
In principle LSTM should be able to learn a smilar rule to Gate. However, our
results do not bear that out.  Understanding
what is going inside the LSTM seems difficult. This is
perhaps another advantage of Gate.

The experimental results show limitations of our approach. Except for Latin,
OCR error rates double compared to the Oracle case on the challenging photo
dataset. We believe that this is explained by the lack of photo training data
for non-Latin languages, as discussed in Section~\ref{sec:results}.
Photos are harder for a number of reasons. Images have text facing
sideways or with deep perspective vanishing point. Characters are
partially occluded, upside down, blurry, or printed on wrinkly clothing.

When script identification fails, text recognition fails. The
characters of one script are not usually supported by another script model.
Even when characters from different scripts look similar
their Unicode codepoints are different. For example, some Cyrillic characters
look similar to Latin.
One way to alleviate this problem is
to consider multiple hypotheses when selecting
models. Eq.~\eqref{eq:ocr-with-scriptid} uses $\max$ to select the best script
and makes a hard decision. Instead, we could try several per-script
OCR models.
This introduces a trade-off between computation effort and accuracy.
Another approach is to design an end-to-end text recognition
system that embeds script identification.
Likely many of the features used
for OCR can be reused for script identification and only be
computed once.

In our approach we take an entire text line as the unit for script
classification. A line image is a natural unit for an OCR system from
a text-detection and layout-analysis point of view.
In addition it presents sufficient
context for the newest sequence to sequence recognizers to deliver
good accuracy by taking into account a language model.
The vast majority of lines encountered many documents and use-cases
will contain only a single language.
This motivated the decision in the prior work
~\cite{DmitriyGenzel:2013:HSI:2505377.2505382}
to aggregate script identification
at the line-level via a counting heuristic.
Nevertheless,
a multilingual OCR system should ideally be able to handle arbitrary
mixed-script and mixed-language content.
One way to deal with this
problem~\cite{DmitriyGenzel:2013:HSI:2505377.2505382,Adnan-Ul-Hasan:2015:ICDAR}
is by detecting scripts at finer levels.
Another way is to keep the line-level approach and consider multiple scripts in
Eq.~\eqref{eq:ocr-with-scriptid}. The post-recognition layout analysis
phase in Fig.~\ref{fig:multilingual-ocr} removing can then remove
overlapping words . It may however be preferable to have a system
that handles multiscript lines directly. 

We addressed the script identification problem by framing it as a
sequence-to-label problem. This formulation is generic and can be applied to
other use cases. For example, line orientation-direction
detection can find the orientation and reading order of
text. Another use case is garbage detection which determines if the input line
is text or not, as a filter for later stages.

\section{Conclusion}
\label{sec:conclusion}

We considered line-level script identification in the context of multilingual
OCR, where a script identification step selects the script-specific model used
by the OCR engine.  We investigated several variants of an
\emph{encoder-summarizer} approach in the context of a state-of-the-art
multilingual OCR system. We used an evaluation set of multiple-domain line
images in 30 scripts from 232 languages. We compared this approach against a
well-performing but expensive prior
baseline~\cite{DmitriyGenzel:2013:HSI:2505377.2505382}.  We found that the best
variants of the new approach improve significantly over the baseline. In
addition the approach offers computational advantage. The most promising
variant is the \emph{Gate} system. It uses learned attention to emphasize
salient features. It reduces script identification errors by 16\%,
and character recognition error rates from script misidentification by a third.

We also discussed a limitation of our approach:
the inability to handle multiple scripts within a line.
Dealing with such cases using a single model is future work.

% The experimental results also show some limitations of our
% approach. It is unable to handle multiple scripts within lines. It has
% a tendency to confuse scripts that can have locally similar or
% identical appearance, for example characters in Cyrillic and Latin
% that look identical. It has somewhat lagging
% accuracy for non-Latin scripts on the photo dataset.

\section*{Acknowledgment}

The authors would like to thank
Thomas Deselaers,
Reeve Ingle,
Sergey Ioffe,
Henry Rowley, and
Ray Smith
for comments on early versions of this paper.



\bibliographystyle{IEEEtran}
\bibliography{IEEEabrv,paper}

% \end{thebibliography}

\begin{table*}
  \centering
\begin{tabular}{|r|p{0.14cm} p{0.14cm} p{0.14cm} p{0.14cm} p{0.14cm} p{0.14cm} p{0.14cm} p{0.14cm} p{0.14cm} p{0.14cm} p{0.14cm} p{0.14cm} p{0.14cm} p{0.14cm} p{0.14cm} p{0.14cm} p{0.14cm} p{0.14cm} p{0.14cm} p{0.14cm} p{0.14cm} p{0.14cm} p{0.14cm} p{0.14cm} p{0.14cm} p{0.14cm} p{0.14cm} p{0.14cm} p{0.14cm} p{0.35cm}|}
\hline
   & \begin{rotate}{270} Arab \end{rotate}
   & \begin{rotate}{270} Armn \end{rotate}
   & \begin{rotate}{270} Beng \end{rotate}
   & \begin{rotate}{270} Cher \end{rotate}
   & \begin{rotate}{270} Cyrl \end{rotate}
   & \begin{rotate}{270} Deva \end{rotate}
   & \begin{rotate}{270} Ethi \end{rotate}
   & \begin{rotate}{270} Geor \end{rotate}
   & \begin{rotate}{270} Grek \end{rotate}
   & \begin{rotate}{270} Gujr \end{rotate}
   & \begin{rotate}{270} Guru \end{rotate}
   & \begin{rotate}{270} Hans \end{rotate}
   & \begin{rotate}{270} Hant \end{rotate}
   & \begin{rotate}{270} Hebr \end{rotate}
   & \begin{rotate}{270} Jpan \end{rotate}
   & \begin{rotate}{270} Khmr \end{rotate}
   & \begin{rotate}{270} Knda \end{rotate}
   & \begin{rotate}{270} Kore \end{rotate}
   & \begin{rotate}{270} Laoo \end{rotate}
   & \begin{rotate}{270} Latn \end{rotate}
   & \begin{rotate}{270} Mlym \end{rotate}
   & \begin{rotate}{270} Mymr \end{rotate}
   & \begin{rotate}{270} Orya \end{rotate}
   & \begin{rotate}{270} Sinh \end{rotate}
   & \begin{rotate}{270} Syrc \end{rotate}
   & \begin{rotate}{270} Taml \end{rotate}
   & \begin{rotate}{270} Telu \end{rotate}
   & \begin{rotate}{270} Thaa \end{rotate}
   & \begin{rotate}{270} Thai \end{rotate}
   & \begin{rotate}{270} Tibt \end{rotate}\\ 
 & & & & & & & & & & & & & & & & & & & & & & & & & & & & & & \\ 
 & & & & & & & & & & & & & & & & & & & & & & & & & & & & & & \\ 
\hline
Arab & 99.1 & 0 & 0 & 0 & 0 & 0 & 0 & 0 & 0 & 0 & 0 & 0 & 0 & 0 & 0 & 0 & 0 & 0 & 0 & 0.8 & 0 & 0 & 0 & 0 & 0 & 0 & 0 & 0 & 0 & 0\\ 
Armn & 0 & 78.8 & 0 & 0 & 0 & 0 & 0 & 0 & 0 & 0 & 0 & 0 & 0 & 0 & 0 & 0 & 0 & 0 & 0 & 21.1 & 0 & 0 & 0 & 0 & 0 & 0 & 0 & 0 & 0 & 0\\ 
Beng & 0 & 0 & 96.8 & 0 & 0 & 0.3 & 0 & 0 & 0 & 0 & 0 & 0 & 0 & 0 & 0 & 0 & 0 & 0 & 0 & 2.8 & 0 & 0 & 0 & 0 & 0 & 0 & 0 & 0 & 0 & 0\\ 
Cher & 0 & 0 & 0 & 94.8 & 0 & 0 & 0 & 0 & 0 & 0 & 0 & 0 & 0 & 0 & 0 & 0 & 0 & 0 & 0 & 5.2 & 0 & 0 & 0 & 0 & 0 & 0 & 0 & 0 & 0 & 0\\ 
Cyrl & 0 & 0 & 0 & 0 & 95.8 & 0 & 0 & 0 & 0 & 0 & 0 & 0 & 0 & 0 & 0 & 0 & 0 & 0 & 0 & 4.2 & 0 & 0 & 0 & 0 & 0 & 0 & 0 & 0 & 0 & 0\\ 
Deva & 0 & 0 & 0 & 0 & 0 & 99.1 & 0 & 0 & 0 & 0 & 0 & 0 & 0 & 0 & 0 & 0 & 0 & 0 & 0 & 0.6 & 0 & 0 & 0 & 0 & 0 & 0 & 0 & 0 & 0 & 0\\ 
Ethi & 0 & 0 & 0 & 0 & 0 & 0 & 99 & 0 & 0 & 0 & 0 & 0 & 0 & 0 & 0 & 0 & 0 & 0 & 0 & 0.9 & 0 & 0 & 0 & 0 & 0 & 0 & 0 & 0 & 0 & 0\\ 
Geor & 0 & 0 & 0 & 0 & 0 & 0 & 0 & 99.3 & 0 & 0 & 0 & 0 & 0 & 0 & 0 & 0 & 0 & 0 & 0 & 0.5 & 0 & 0 & 0 & 0 & 0 & 0 & 0 & 0 & 0 & 0\\ 
Grek & 0 & 0 & 0 & 0 & 0.1 & 0 & 0 & 0 & 87.6 & 0 & 0 & 0 & 0 & 0 & 0 & 0 & 0 & 0 & 0 & 12.3 & 0 & 0 & 0 & 0 & 0 & 0 & 0 & 0 & 0 & 0\\ 
Gujr & 0 & 0 & 0 & 0 & 0 & 0.3 & 0 & 0 & 0 & 99 & 0 & 0 & 0 & 0 & 0 & 0 & 0 & 0 & 0 & 0.8 & 0 & 0 & 0 & 0 & 0 & 0 & 0 & 0 & 0 & 0\\ 
Guru & 2.4 & 0 & 0 & 0 & 0 & 0.5 & 0 & 0 & 0 & 0 & 95.3 & 0 & 0 & 0 & 0 & 0 & 0 & 0 & 0 & 1.8 & 0 & 0 & 0 & 0 & 0 & 0 & 0 & 0 & 0 & 0\\ 
Hans & 0 & 0 & 0 & 0 & 0 & 0 & 0 & 0 & 0 & 0 & 0 & 98.2 & 1.7 & 0 & 0 & 0 & 0 & 0 & 0 & 0.1 & 0 & 0 & 0 & 0 & 0 & 0 & 0 & 0 & 0 & 0\\ 
Hant & 0 & 0 & 0 & 0 & 0 & 0 & 0 & 0 & 0 & 0 & 0 & 7.3 & 92 & 0 & 0.6 & 0 & 0 & 0 & 0 & 0 & 0 & 0 & 0 & 0 & 0 & 0 & 0 & 0 & 0 & 0\\ 
Hebr & 0 & 0 & 0 & 0 & 0 & 0 & 0 & 0 & 0 & 0 & 0 & 0 & 0 & 99.8 & 0 & 0 & 0 & 0 & 0 & 0.2 & 0 & 0 & 0 & 0 & 0 & 0 & 0 & 0 & 0 & 0\\ 
Jpan & 0 & 0 & 0 & 0 & 0 & 0 & 0 & 0 & 0 & 0 & 0 & 1.9 & 2.6 & 0 & 95.3 & 0 & 0 & 0 & 0 & 0.2 & 0 & 0 & 0 & 0 & 0 & 0 & 0 & 0 & 0 & 0\\ 
Khmr & 0 & 0 & 0 & 0 & 0 & 0 & 0 & 0 & 0 & 0 & 0 & 0.2 & 0 & 0 & 0 & 81.4 & 0 & 0 & 0 & 15.7 & 0 & 0 & 0 & 0 & 0 & 0 & 0 & 0 & 2.6 & 0\\ 
Knda & 0 & 0 & 0 & 0 & 0 & 0 & 0 & 0 & 0 & 0 & 0 & 0 & 0 & 0 & 0 & 0 & 99 & 0 & 0 & 0.4 & 0 & 0 & 0 & 0 & 0 & 0 & 0.5 & 0 & 0 & 0\\ 
Kore & 0.1 & 0 & 0 & 0 & 0 & 0 & 0 & 0 & 0.1 & 0 & 0 & 0.1 & 1.3 & 0 & 0 & 0 & 0 & 97.9 & 0 & 0.5 & 0 & 0 & 0 & 0 & 0 & 0 & 0 & 0 & 0 & 0\\ 
Laoo & 0 & 0 & 0 & 0.2 & 0 & 0 & 0 & 0 & 0 & 0 & 0 & 0 & 0 & 0 & 0 & 0.6 & 0 & 0 & 93.3 & 3.5 & 0 & 0.8 & 0 & 0 & 0 & 0 & 0.6 & 0 & 0.9 & 0\\ 
Latn & 0 & 0 & 0 & 0 & 0 & 0 & 0 & 0 & 0 & 0 & 0 & 0 & 0 & 0 & 0 & 0 & 0 & 0 & 0 & 99.7 & 0 & 0 & 0 & 0 & 0 & 0 & 0 & 0 & 0 & 0\\ 
Mlym & 0 & 0 & 0 & 0 & 0 & 0 & 0 & 0 & 0 & 0 & 0 & 0 & 0 & 0 & 0 & 0 & 0 & 0 & 0 & 1.2 & 98.8 & 0 & 0 & 0 & 0 & 0 & 0 & 0 & 0 & 0\\ 
Mymr & 0 & 0 & 0 & 0 & 0 & 0 & 0 & 0 & 0 & 0 & 0 & 0 & 0 & 0 & 0 & 0 & 0 & 0 & 0 & 1 & 0 & 98.5 & 0 & 0 & 0 & 0 & 0.1 & 0 & 0 & 0\\ 
Orya & 0 & 0 & 0.1 & 0 & 0 & 0 & 0 & 0 & 0 & 0 & 0 & 0 & 0 & 0 & 0 & 0 & 0 & 0 & 0 & 4.4 & 0 & 0 & 95 & 0 & 0 & 0 & 0 & 0 & 0 & 0\\ 
Sinh & 0 & 0 & 0 & 0 & 0 & 0 & 0 & 0 & 0 & 0 & 0 & 0 & 0 & 0 & 0 & 0 & 0.2 & 0 & 0 & 0.7 & 0 & 0 & 0 & 98.8 & 0 & 0 & 0.1 & 0 & 0 & 0\\ 
Syrc & 1.2 & 0 & 0 & 0 & 0 & 0 & 0 & 0 & 0 & 0 & 0 & 0 & 0 & 0 & 0 & 0 & 0 & 0 & 0 & 2.9 & 0 & 0 & 0 & 0 & 95.6 & 0 & 0 & 0 & 0 & 0\\ 
Taml & 0 & 0 & 0 & 0 & 0 & 0 & 0 & 0 & 0 & 0 & 0 & 0 & 0 & 0 & 0 & 0 & 0 & 0 & 0 & 3.4 & 0 & 0 & 0 & 0 & 0 & 96.6 & 0 & 0 & 0 & 0\\ 
Telu & 0 & 0 & 0 & 0 & 0 & 0 & 0 & 0 & 0 & 0 & 0 & 0 & 0 & 0 & 0 & 0 & 0 & 0 & 0 & 1 & 0 & 0 & 0 & 0 & 0 & 0 & 99 & 0 & 0 & 0\\ 
Thaa & 0.2 & 0 & 0 & 0 & 0 & 0 & 0 & 0 & 0 & 0 & 0 & 0 & 0 & 0 & 0 & 0 & 0 & 0 & 0 & 0.5 & 0 & 0 & 0 & 0 & 0 & 0 & 0 & 99.3 & 0 & 0\\ 
Thai & 0 & 0 & 0 & 0 & 0 & 0 & 0 & 0 & 0 & 0 & 0 & 0 & 0 & 0 & 0 & 0 & 0 & 0 & 0 & 0.3 & 0 & 0 & 0 & 0 & 0 & 0 & 0 & 0 & 99.7 & 0\\ 
Tibt & 0.1 & 0 & 0 & 0 & 0 & 0.1 & 0 & 0 & 0 & 0 & 0.3 & 0 & 0 & 0 & 0 & 0 & 0 & 0 & 0 & 1.3 & 0 & 0 & 0 & 0 & 0.2 & 0 & 0 & 0 & 0 & 97.6\\
\hline\end{tabular}

  \caption{Full confusion matrix for Base }
  \label{tab:full_confusion_base}
\end{table*}

\begin{table*}
  \centering
\begin{tabular}{|r|p{0.14cm} p{0.14cm} p{0.14cm} p{0.14cm} p{0.14cm} p{0.14cm} p{0.14cm} p{0.14cm} p{0.14cm} p{0.14cm} p{0.14cm} p{0.14cm} p{0.14cm} p{0.14cm} p{0.14cm} p{0.14cm} p{0.14cm} p{0.14cm} p{0.14cm} p{0.14cm} p{0.14cm} p{0.14cm} p{0.14cm} p{0.14cm} p{0.14cm} p{0.14cm} p{0.14cm} p{0.14cm} p{0.14cm} p{0.35cm}|}
\hline
   & \begin{rotate}{270} Arab \end{rotate}
   & \begin{rotate}{270} Armn \end{rotate}
   & \begin{rotate}{270} Beng \end{rotate}
   & \begin{rotate}{270} Cher \end{rotate}
   & \begin{rotate}{270} Cyrl \end{rotate}
   & \begin{rotate}{270} Deva \end{rotate}
   & \begin{rotate}{270} Ethi \end{rotate}
   & \begin{rotate}{270} Geor \end{rotate}
   & \begin{rotate}{270} Grek \end{rotate}
   & \begin{rotate}{270} Gujr \end{rotate}
   & \begin{rotate}{270} Guru \end{rotate}
   & \begin{rotate}{270} Hans \end{rotate}
   & \begin{rotate}{270} Hant \end{rotate}
   & \begin{rotate}{270} Hebr \end{rotate}
   & \begin{rotate}{270} Jpan \end{rotate}
   & \begin{rotate}{270} Khmr \end{rotate}
   & \begin{rotate}{270} Knda \end{rotate}
   & \begin{rotate}{270} Kore \end{rotate}
   & \begin{rotate}{270} Laoo \end{rotate}
   & \begin{rotate}{270} Latn \end{rotate}
   & \begin{rotate}{270} Mlym \end{rotate}
   & \begin{rotate}{270} Mymr \end{rotate}
   & \begin{rotate}{270} Orya \end{rotate}
   & \begin{rotate}{270} Sinh \end{rotate}
   & \begin{rotate}{270} Syrc \end{rotate}
   & \begin{rotate}{270} Taml \end{rotate}
   & \begin{rotate}{270} Telu \end{rotate}
   & \begin{rotate}{270} Thaa \end{rotate}
   & \begin{rotate}{270} Thai \end{rotate}
   & \begin{rotate}{270} Tibt \end{rotate}\\ 
 & & & & & & & & & & & & & & & & & & & & & & & & & & & & & & \\ 
 & & & & & & & & & & & & & & & & & & & & & & & & & & & & & & \\ 
\hline
Arab & 98.9 & 0 & 0 & 0 & 0 & 0 & 0 & 0 & 0 & 0 & 0 & 0 & 0 & 0 & 0 & 0 & 0 & 0 & 0 & 0.4 & 0 & 0 & 0 & 0 & 0.1 & 0 & 0 & 0 & 0 & 0\\ 
Armn & 0 & 97.4 & 0 & 0 & 0 & 0 & 0 & 0 & 0 & 0 & 0 & 0 & 0 & 0 & 0 & 0 & 0 & 0 & 0 & 2.3 & 0 & 0 & 0 & 0 & 0 & 0 & 0 & 0 & 0 & 0\\ 
Beng & 0 & 0 & 97 & 0 & 0 & 0.6 & 0 & 0 & 0 & 0 & 0.2 & 0 & 0 & 0 & 0 & 0 & 0 & 0 & 0 & 1.6 & 0 & 0 & 0 & 0 & 0.1 & 0 & 0 & 0 & 0 & 0\\ 
Cher & 0 & 0 & 0 & 96.3 & 0.2 & 0 & 0.1 & 0 & 0 & 0 & 0 & 0 & 0 & 0 & 0 & 0 & 0 & 0 & 0 & 2.6 & 0 & 0 & 0 & 0 & 0 & 0 & 0 & 0 & 0.1 & 0\\ 
Cyrl & 0 & 0 & 0 & 0 & 98 & 0 & 0 & 0 & 0 & 0 & 0 & 0 & 0 & 0 & 0 & 0 & 0 & 0 & 0 & 1.8 & 0 & 0 & 0 & 0 & 0 & 0 & 0 & 0 & 0 & 0\\ 
Deva & 0 & 0 & 0 & 0 & 0 & 99 & 0 & 0 & 0 & 0 & 0.2 & 0.1 & 0 & 0 & 0 & 0 & 0 & 0 & 0 & 0.2 & 0 & 0 & 0 & 0 & 0 & 0 & 0 & 0 & 0 & 0\\ 
Ethi & 0 & 0 & 0 & 0 & 0 & 0 & 98.6 & 0 & 0 & 0 & 0 & 0.2 & 0 & 0 & 0 & 0 & 0 & 0 & 0 & 0.7 & 0 & 0 & 0 & 0 & 0 & 0 & 0 & 0 & 0 & 0\\ 
Geor & 0 & 0 & 0 & 0 & 0 & 0 & 0 & 99 & 0 & 0 & 0 & 0 & 0 & 0 & 0 & 0 & 0 & 0 & 0 & 0.5 & 0 & 0 & 0 & 0 & 0 & 0 & 0 & 0 & 0 & 0\\ 
Grek & 0 & 0 & 0 & 0 & 0 & 0 & 0 & 0 & 98.1 & 0 & 0 & 0 & 0 & 0 & 0 & 0 & 0 & 0 & 0 & 1.4 & 0 & 0 & 0 & 0 & 0 & 0 & 0 & 0 & 0 & 0\\ 
Gujr & 0 & 0 & 0 & 0 & 0 & 0.3 & 0 & 0 & 0 & 99.6 & 0 & 0 & 0 & 0 & 0 & 0 & 0 & 0 & 0 & 0.2 & 0 & 0 & 0 & 0 & 0 & 0 & 0 & 0 & 0 & 0\\ 
Guru & 2.5 & 0 & 0 & 0 & 0 & 0.6 & 0 & 0 & 0 & 0 & 95.7 & 0 & 0 & 0 & 0 & 0 & 0 & 0 & 0 & 1.1 & 0 & 0 & 0 & 0 & 0 & 0 & 0 & 0 & 0 & 0\\ 
Hans & 0 & 0 & 0 & 0 & 0.2 & 0 & 0 & 0 & 0 & 0 & 0 & 95 & 3.7 & 0 & 0.8 & 0 & 0 & 0 & 0 & 0.2 & 0 & 0 & 0 & 0 & 0 & 0 & 0 & 0 & 0 & 0\\ 
Hant & 0 & 0 & 0 & 0 & 0 & 0 & 0 & 0 & 0 & 0 & 0 & 0.8 & 99.1 & 0 & 0 & 0 & 0 & 0 & 0 & 0 & 0 & 0 & 0 & 0 & 0 & 0 & 0 & 0 & 0 & 0\\ 
Hebr & 0 & 0 & 0 & 0 & 0 & 0 & 0 & 0 & 0 & 0 & 0 & 0 & 0 & 99.9 & 0 & 0 & 0 & 0 & 0 & 0 & 0 & 0 & 0 & 0 & 0 & 0 & 0 & 0 & 0 & 0\\ 
Jpan & 0 & 0 & 0 & 0 & 0 & 0 & 0 & 0 & 0 & 0 & 0 & 0.6 & 3.9 & 0 & 95.4 & 0 & 0 & 0 & 0 & 0 & 0 & 0 & 0 & 0 & 0 & 0 & 0 & 0 & 0 & 0\\ 
Khmr & 0 & 0 & 0 & 0 & 0 & 0 & 0.2 & 0 & 0 & 0 & 0 & 0 & 0 & 0 & 0 & 92.1 & 0.2 & 0 & 1.7 & 5.5 & 0 & 0 & 0 & 0 & 0 & 0 & 0 & 0 & 0.3 & 0\\ 
Knda & 0 & 0 & 0 & 0 & 0 & 0 & 0 & 0 & 0 & 0 & 0 & 0 & 0 & 0 & 0 & 0 & 99.3 & 0 & 0 & 0 & 0 & 0 & 0 & 0 & 0 & 0 & 0.6 & 0 & 0 & 0\\ 
Kore & 0 & 0 & 0 & 0 & 0 & 0 & 0 & 0 & 0 & 0 & 0.2 & 0.3 & 1.2 & 0 & 0 & 0 & 0.1 & 97.7 & 0 & 0.3 & 0 & 0 & 0 & 0 & 0 & 0 & 0 & 0 & 0 & 0\\ 
Laoo & 0 & 0 & 0 & 0 & 0 & 0 & 0 & 0 & 0 & 0 & 0 & 0 & 0 & 0.2 & 0 & 0.7 & 0 & 0 & 95.4 & 1.6 & 0 & 0.4 & 0 & 0 & 0 & 0 & 1.4 & 0 & 0.3 & 0\\ 
Latn & 0 & 0 & 0 & 0 & 0.4 & 0 & 0 & 0 & 0 & 0 & 0 & 0 & 0 & 0 & 0 & 0 & 0 & 0 & 0 & 98.8 & 0 & 0 & 0 & 0 & 0 & 0 & 0 & 0 & 0 & 0\\ 
Mlym & 0 & 0 & 0 & 0 & 0 & 0 & 0 & 0 & 0 & 0 & 0 & 0 & 0 & 0 & 0 & 0 & 0 & 0 & 0 & 0 & 99.9 & 0 & 0 & 0 & 0 & 0 & 0 & 0 & 0 & 0\\ 
Mymr & 0 & 0 & 0 & 0 & 0.1 & 0 & 0 & 0 & 0 & 0 & 0 & 0 & 0 & 0 & 0 & 0 & 0 & 0 & 0.1 & 0.2 & 0.2 & 98.4 & 0.1 & 0 & 0 & 0 & 0.2 & 0 & 0 & 0\\ 
Orya & 0 & 0 & 0 & 0 & 0.3 & 0 & 0 & 0 & 0 & 0 & 0.2 & 0 & 0.1 & 0 & 0 & 0.4 & 0 & 0 & 0 & 2.2 & 0 & 0.2 & 96 & 0 & 0 & 0 & 0 & 0 & 0 & 0\\ 
Sinh & 0 & 0 & 0 & 0 & 0 & 0 & 0 & 0 & 0 & 0 & 0.2 & 0.1 & 0 & 0 & 0 & 0 & 0.1 & 0 & 0 & 0.6 & 0 & 0 & 0.1 & 98.4 & 0 & 0 & 0 & 0 & 0 & 0\\ 
Syrc & 0.9 & 0 & 0 & 0 & 0 & 0 & 0 & 0 & 0 & 0 & 0.3 & 0 & 0 & 0.2 & 0 & 0 & 0 & 0 & 0 & 1.2 & 0.1 & 0 & 0 & 0 & 95.4 & 0 & 0.1 & 0 & 0 & 0.3\\ 
Taml & 0.1 & 0 & 0 & 0 & 0 & 0 & 0 & 0 & 0 & 0 & 0 & 0 & 0 & 0 & 0 & 0 & 0 & 0 & 0 & 0.3 & 0 & 0 & 0 & 0 & 0 & 99.6 & 0 & 0 & 0 & 0\\ 
Telu & 0 & 0 & 0 & 0 & 0 & 0 & 0.2 & 0 & 0 & 0 & 0 & 0 & 0 & 0 & 0 & 0 & 0 & 0 & 0 & 0.3 & 0 & 0 & 0 & 0 & 0 & 0 & 99.6 & 0 & 0 & 0\\ 
Thaa & 0 & 0 & 0 & 0 & 0 & 0 & 0 & 0 & 0 & 0 & 0 & 0 & 0 & 0 & 0 & 0 & 0 & 0 & 0 & 0 & 0 & 0 & 0 & 0 & 0 & 0 & 0 & 99.4 & 0 & 0\\ 
Thai & 0 & 0 & 0 & 0 & 0 & 0 & 0 & 0 & 0 & 0 & 0 & 0 & 0 & 0 & 0 & 0 & 0 & 0 & 0.2 & 0.2 & 0 & 0 & 0 & 0 & 0 & 0 & 0 & 0 & 99.6 & 0\\ 
Tibt & 0 & 0 & 0 & 0 & 0 & 0.3 & 0 & 0 & 0 & 0 & 0.5 & 0 & 0 & 0 & 0 & 0.1 & 0 & 0 & 0 & 0.3 & 0 & 0 & 0 & 0 & 0.2 & 0 & 0 & 0 & 0 & 97.1\\
\hline\end{tabular}

  \caption{Full confusion matrix for Gate }
  \label{tab:full_confusion_gate}
\end{table*}

\end{document}